\documentclass[a4paper, fleqn]{cas-sc}
\usepackage[numbers]{natbib}
\usepackage{textcomp}
\usepackage{url}
\usepackage{verbatim}
\usepackage{graphicx}
\usepackage{amssymb}
\usepackage{amsmath}
\usepackage{array,booktabs}
\usepackage{multirow}
\usepackage{bm}
\usepackage{bbding}
\usepackage{pifont}
\usepackage{makecell}
\usepackage{xcolor}
\usepackage{array}
\usepackage{tabularx}
\usepackage{newfloat}
\usepackage{listings}
\usepackage{algorithm}
\usepackage{algorithmic}
\usepackage{hyperref}
\usepackage{subcaption}
\usepackage{multicol}

\def\tsc#1{\csdef{#1}{\textsc{\lowercase{#1}}\xspace}}
\tsc{WGM}
\tsc{QE}
\tsc{EP}
\tsc{PMS}
\tsc{BEC}
\tsc{DE}
\begin{document}
\begin{sloppypar}
\let\WriteBookmarks\relax
\def\floatpagepagefraction{1}
\def\textpagefraction{.001}
\let\printorcid\relax
\shorttitle{}

\shortauthors{Liao et~al.}

\title [mode = title]{Addressing Corner Cases in Autonomous Driving: A World Model-based Approach with Mixture of Experts and LLMs}                      
\author[1,2]{Haicheng Liao}
\fnmark[1]
\credit{Conceptualization, Methodology, Writing}

\author[1,2]{Bonan Wang}
\credit{Experiment, Writing}
\fnmark[1]

\author[1]{Junxian Yang}
\credit{Experiment, Writing}

\author[1,3]{Chengyue Wang}
\credit{Methodology}

\author[4]{Zhengbing He}
\credit{Methodology}

\author[5]{Guohui Zhang}
\credit{Methodology}

\author[1,2]{Chengzhong Xu}

\credit{Conceptualization, Funding, Review}

\author[1,2,3]{Zhenning Li}
\cormark[1]
\ead{zhenningli@um.edu.mo}
\credit{Conceptualization}

\affiliation[1]{organization={State Key Laboratory of Internet of Things for Smart City, University of Macau}, city={Macau SAR}, country={China}}

\affiliation[2]{organization={Department of Computer and Information Science, University of Macau}, city={Macau SAR}, country={China}}

\affiliation[3]{organization={Department of Civil and Environmental Engineering, University of Macau}, city={Macau SAR}, country={China}}

\affiliation[4]{organization={Senseable City Lab, Massachusetts Institute of Technology}, city={Cambridge MA}, country={United States}}

\affiliation[5]{organization={Department of Civil and Environmental Engineering, University of Hawaii}, city={Honolulu HI}, country={United States}}

\cortext[cor1]{Corresponding author; $^{1}$Equally Contributed}

\begin{abstract}
Accurate and reliable motion forecasting is essential for the safe deployment of autonomous vehicles (AVs), particularly in rare but safety-critical scenarios known as corner cases. Existing models often underperform in these situations due to an over-representation of common scenes in training data and limited generalization capabilities. To address this limitation, we present WM-MoE, the first world model-based motion forecasting framework that unifies perception, temporal memory, and decision making to address the challenges of high-risk corner-case scenarios. The model constructs a compact scene representation that explains current observations, anticipates future dynamics, and evaluates the outcomes of potential actions. To enhance long-horizon reasoning, we leverage large language models (LLMs) and introduce a lightweight temporal tokenizer that maps agent trajectories and contextual cues into the LLM’s feature space without additional training, enriching temporal context and commonsense priors. Furthermore, a mixture-of-experts (MoE) is introduced to decompose complex corner cases into subproblems and allocate capacity across scenario types, and a router assigns scenes to specialized experts that infer agent intent and perform counterfactual rollouts. In addition, we introduce nuScenes-corner, a new benchmark that comprises four real-world corner-case scenarios for rigorous evaluation.
Extensive experiments on four benchmark datasets (nuScenes, NGSIM, HighD, and MoCAD) showcase that WM-MoE consistently outperforms state-of-the-art (SOTA) baselines and remains robust under corner-case and data-missing conditions, indicating the promise of world model–based architectures for robust and generalizable motion forecasting in fully AVs.
\end{abstract}

\begin{keywords}
Autonomous Driving \sep World Models \sep Motion Forecasting  \sep Large Language Models \sep Mixture of Experts Models

\end{keywords}

\maketitle

\section{Introduction}
The development of autonomous vehicles (AVs) has the potential to revolutionize transportation, offering significant advancements in safety, efficiency, and convenience \cite{long2025physical,wang2025beyond}. A critical component of AV technology is the ability to accurately predict the future trajectories of surrounding agents, such as vehicles, pedestrians, and cyclists. Despite the progress in this field, existing motion forecasting models often struggle with ``corner cases''—scenes characterized by high uncertainty, rare events, and complex interactions among multiple agents  \cite{sheng2024ego,zhang2025mm}.

\begin{figure}
  \centering
\includegraphics[width=0.65\linewidth]{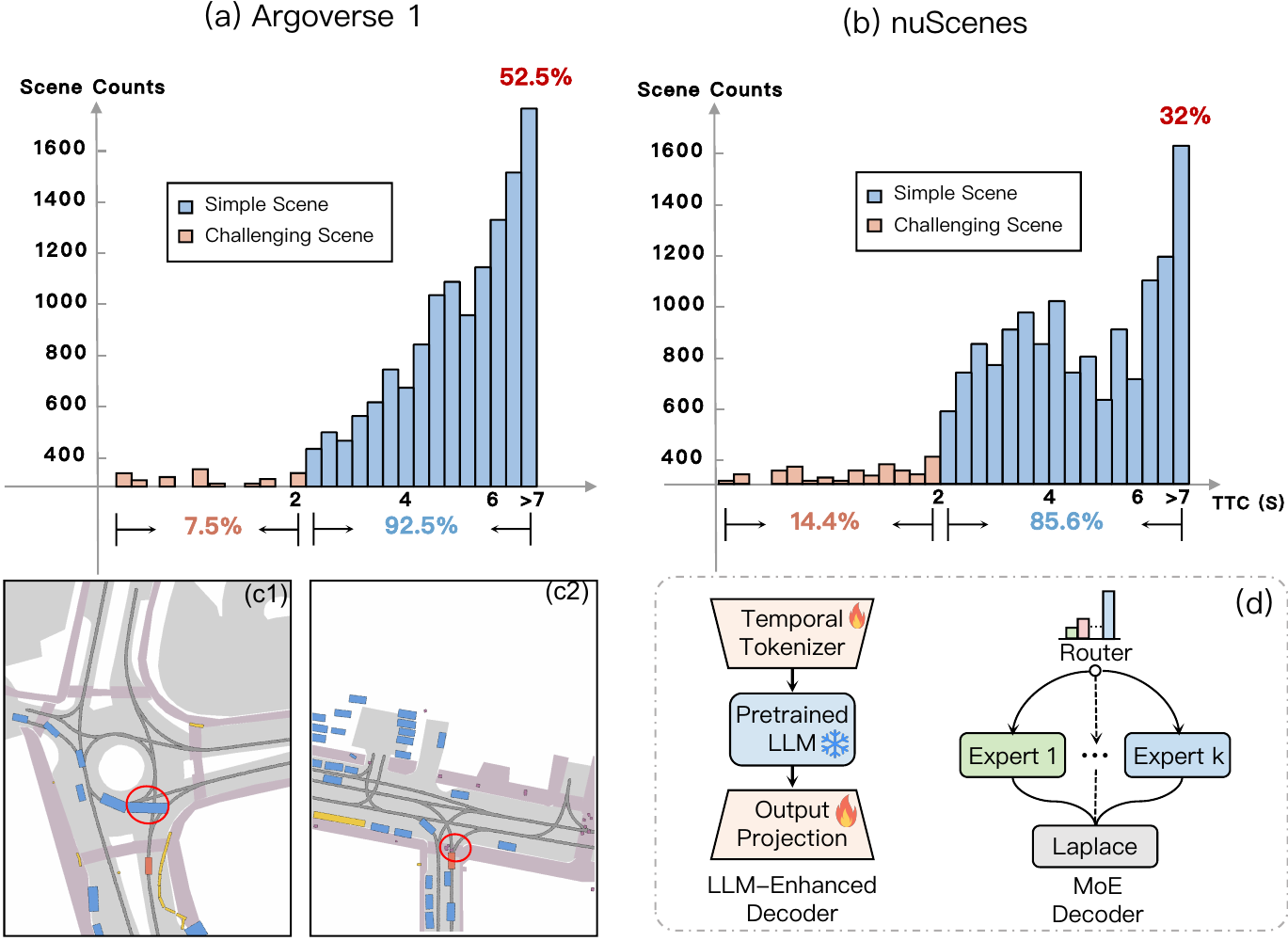}
  \caption{Skewed time to collision (TTC) distributions and key highlights of our proposed model. (a) and (b) illustrate the skewed distributions of TTC in the nuScenes and Argoverse 1 datasets, with a small percentage of high-risk scenarios (TTC $<$ 2s). (c) highlights examples of safety-critical situations in the nuScenes dataset, like turns and roundabouts, while other agents are around. (d) presents the key highlight of our work: the integration of a MoE model and LLMs with the proposed temporal tokenizer to enhance performance in challenging scenarios while maintaining accuracy in simpler ones.}
  \label{head}
\end{figure}

In essence, real-world scenarios are inherently imbalanced: common and straightforward traffic agent movements coexist with relatively rare but challenging complex corner cases. These corner cases, though infrequent, are among the most critical scenarios that an autonomous driving (AD) system must manage effectively. Such cases include sudden lane changes, abrupt stops, unexpected pedestrian movements, and intricate maneuvers at intersections \cite{jia2025cltp}. Traditional motion forecasting models \cite{liao2024bat,liao2025minds} are predominantly trained on datasets composed primarily of common driving scenarios, where vehicles adhere to straightforward kinematic rules. Due to their higher frequency and lower complexity, deep learning-based models are biased toward handling these scenarios during training. In contrast, corner-case scenes are underrepresented and receive less attention from these models. Consequently, these models tend to perform poorly in rare but safety-critical corner cases, leading to potentially catastrophic outcomes when AVs encounter such high-risk situations, compromising safety and reliability precisely when they are needed most \cite{liao2025sa}.

{ 
To address this challenge, early research \cite{ouyang2016factors} efforts involve oversampling corner-case scenarios or undersampling more frequent ones to mitigate model imbalance. Additionally, some approaches \cite{ross2017focal,tao2023local} reweight the loss function to amplify the significance of corner-case samples. Another solution \cite{liu2022open} is to categorize scenarios and apply classification learning, where increasing the feature distance between different scenario categories and reducing it within the same category encouraged models to learn the feature space of rare scenarios, improving generalization. Although these methods have shown some effectiveness, they often improve corner-case performance at the cost of degrading the model's accuracy on more frequent scenarios.
As a step forward from imbalanced learning, contrastive learning has recently attracted significant research interest as a promising method for addressing uneven data distribution. By distinguishing positive samples from others, contrastive learning purposefully organizes the latent feature space to achieve more balanced training. However, contrastive learning methods \cite{lan2024hi} introduce substantial computational overhead, negatively impacting prediction efficiency. More importantly, these methods typically cannot explicitly classify different scenario categories, instead relying on implicit differentiation. This implicit classification limits the model's ability to effectively distinguish features across diverse scenarios, potentially resulting in insufficient modeling of rare but critical corner cases. Moreover, the lack of explicit category distinction reduces the model's interpretability and debuggability, thereby hindering its generalization capability in complex traffic scenes \cite{dai2025large}.

Given the limitations of both data balancing and representation learning approaches, researchers have also explored imitation learning (IL) as a pathway to achieve more human-like motion forecasting. IL-based models \cite{liao2025toward,wang2025nest} aim to produce human-like actions, but they often suffer from distribution shift—a mismatch between training (open-loop) and real-world execution (closed-loop). In practice, small prediction errors can compound over time, especially in rare or complex scenes, leading to unsafe behavior. Furthermore, IL models rely solely on observational data, which often encourages learning shallow correlations rather than true causal reasoning. This makes them brittle in out-of-distribution (OOD) and long-tail scenarios. Figure \ref{head} provides a clear illustration of this imbalance by showing the distribution of TTC between pairs of vehicles in the nuScenes and Argoverse 1 datasets. Here, scenarios with a TTC of less than 2 seconds are classified as high-risk, representing safety-critical situations that demand precise motion forecasting. Our observations reveal a pronounced skew in the distribution of TTCs, with a majority of scenarios falling into low-risk categories and a minority representing the high-risk, corner cases. This skew highlights the challenge that current models face: they are not adequately equipped to deal with the scenarios that are most likely to result in safety-critical failures.
For safe navigation, AD systems must be trained on vast and diverse datasets that adequately represent even the rarest scenarios. However, even with extensive data and simulated testing, current methods still fall short of human-level reliability in real-world driving. 
These challenges prompt a critical question: How can we combine complementary paradigms to handle corner cases and generalize beyond the training distribution?

In practice, human drivers navigate diverse and unfamiliar situations with ease. This capability stems from the human brain’s ``world model'': an internal understanding of how the world evolves \cite{guan2024world}. Specifically, human drivers build a compact, spatiotemporal abstract representation of the physical world from limited and noisy sensory input. This abstract representation makes the scene easier to predict and supports the acquisition of background knowledge about how the world works. With such knowledge, a human driver is able to decompose complex maneuvers into simpler, lower-level actions and apply them flexibly across unseen situations, and simulate plausible future states to update the world model and make decisions accordingly.

\begin{figure}
  \centering
\includegraphics[width=0.95\linewidth]{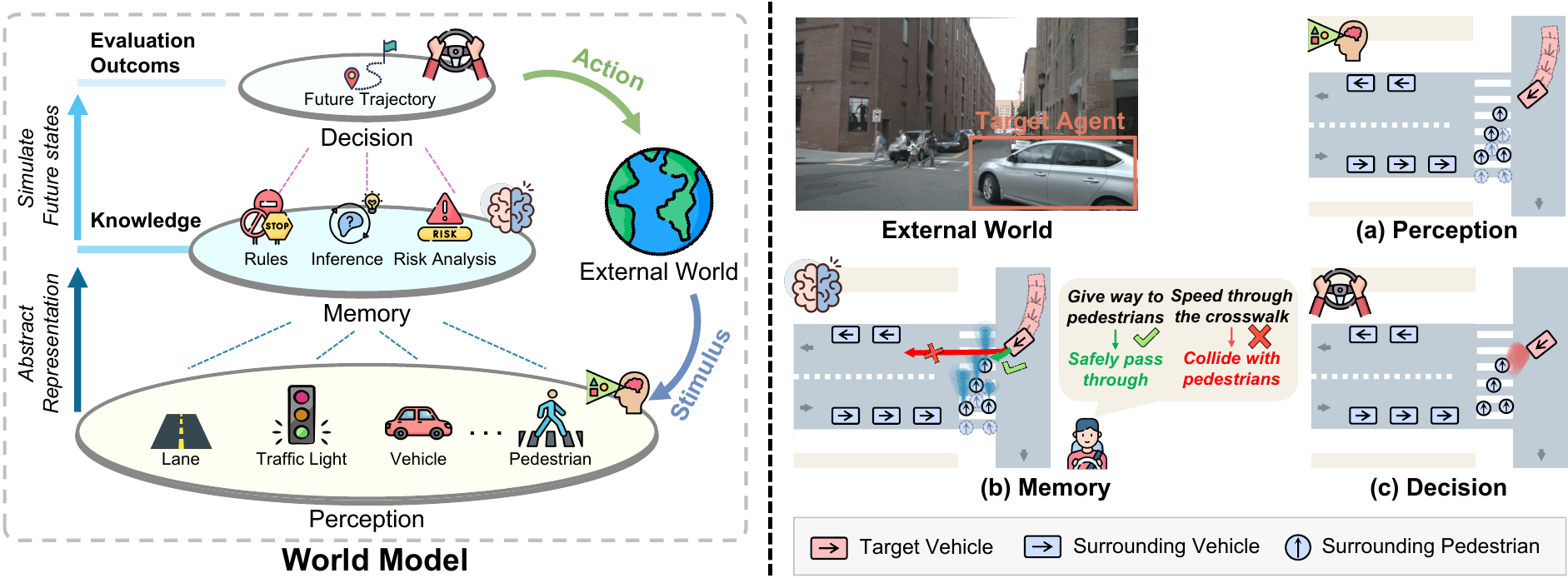}
  \caption{{ Illustration of the world model guiding driving decisions. Left: The world model maintains a compact latent scene state via three interacting modules: Perception, Memory, and Decision. Right: Imagine a corner-case scenario involving a partially occluded crosswalk with pedestrians entering. The human brain's internal world model navigates this unseen scene through a sequence of operations: (a) First, the Perception module abstracts the complex external world into a simplified, task-relevant that localizes the target agent, nearby pedestrians, and road geometry; (b) Next, the Memory module contextualizes this sketch, integrating recent history with commonsense physical and social rules (e.g., ``yield to pedestrians''), enabling counterfactual reasoning that contrasts a safe yield with an unsafe acceleration. (c) Finally, the Decision module evaluates these simulated futures and selects the optimal action, resulting in a safe and rational maneuver.}}

  \label{head_2}
\end{figure}

As shown in Figure \ref{head_2}, we can roughly view the driver’s internal world model as a compact scene state that supports three core functions \cite{guan2024world}: (i) explains current and historical observations, (ii) predicts how the scene evolves, and (iii) estimates the sensory consequences of possible actions. Consider a corner-case scenario at a crosswalk where the view is partially obstructed and pedestrians begin to cross. The human driver's perception module first abstracts the complex external world into a simplified, task-relevant sketch, identifying the target agents, nearby pedestrians, and the road layout. Guided by this simplified sketch of the external world, the memory module invokes commonsense physical and social constraints like ``yield to pedestrians'', prioritizing unexpected cues that may indicate danger or call for adaptation. Then, the internal world model of the driver's brain engages in counterfactual reasoning, simulating two possible futures: one where the driver yields, resulting in a safe passage, and another where the driver speeds through, leading to a collision. Finally, the decision module evaluates the outcomes of these simulated futures, leading to the rational choice to decelerate and grant the pedestrian the right-of-way. By supporting counterfactual reasoning about how events might unfold, this internal scene state enables drivers to learn structured representations from prior experience and present context, thereby facilitating safety-critical decisions.

Building on this intuition, we are led to a pivotal research question: \emph{can AV motion forecasting benefit from an explicit world model while giving special treatment to corner-case, high-risk situations?} To address this, we introduce WM-MoE, a world-model–based mixture-of-experts framework that explicitly distinguishes imbalanced scenario types and allocates capacity to rare, high-risk corner cases, while maintaining accuracy on frequent scenes.
Our WM-MoE follows human brain-inspired structure and comprises three components: (1) a perception module that compresses high-dimensional sensor input into a concise scene representation; (2) a memory module that maintains and updates temporal context to predict how this representation will evolve; and (3) a decision module that performs prediction by running counterfactual rollouts over latent states and selecting plausible futures consistent with scene context.

To improve long-term reasoning, we use large language models (LLMs) with excellent multimodal fusion and temporal capabilities to enhance scene comprehension and contextual reasoning, thereby facilitating the extraction of high-dimensional representations of the physical world. Moreover, we also introduce a temporal tokenizer that maps time series into the LLM’s multimodal feature space without further LLM training. This improves temporal coherence, injects commonsense context, and helps the memory module anticipate scene evolution driven by interacting agents. Importantly, in the decision module, we introduce a Mixture-of-Experts (MoE) to specialize processing across scenario types: lightweight routers assign each scene to expert subnetworks that infer other agents’ intentions and simulate plausible future outcomes. By dedicating experts to rare corner cases, the model learns subtle, safety-critical patterns and performs human-like, ``what-if’’ counterfactual reasoning to produce reliable final predictions.  

In summary, the main contributions of this study are as follows:

\begin{itemize}
\item We propose the first \textbf{world model-based} motion forecasting framework that integrates the MoE network with LLMs. A lightweight temporal tokenizer maps trajectories and context into an LLM feature space without additional training, enriching the world model’s temporal memory and priors to yield structured latent representations. Within this latent space, the MoE specializes by routing scenes to different experts that learn dynamic patterns of agent interaction, enabling proactive, counterfactual reasoning over plausible future motions.

\item We develop a new dataset, \textbf{nuScenes-corner}, specifically crafted to advance motion forecasting research. This dataset comprises rare and imbalanced scenarios extracted from the nuScenes dataset, providing a comprehensive benchmark for evaluating the performance of motion forecasting models in challenging corner-case scenes.

\item Our proposed model surpasses the top baseline models on the four real-world datasets: nuScenes, NGSIM, HighD, and MoCAD. Importantly, it maintains top performance in challenging scenarios, including \textbf{corner-case} and \textbf{data-missing} conditions, across diverse environments such as highways, intersections, and urban areas.

\end{itemize}
The paper is structured as follows: Section \ref{Related Work} reviews related literature. Section \ref{Method} presents the motion forecasting problem and describes the components of our proposed model. Section \ref{Experiment} evaluates the model's performance on four real-world datasets, including extensive ablation and qualitative analyses. Finally, Section \ref{Conclusion} concludes the paper with a summary of our findings and reflections on this study.}

\section{Related Work}\label{Related Work}{ 

{ 
\subsection{World Models in Autonomous Driving}
World models have recently attracted broad attention for their potential to narrow the gap between human and machine intelligence in autonomous driving \cite{guan2024world}. A world model learns a general representation of the environment and predicts future states given a sequence of actions. This capability allows systems to anticipate change and adapt their behavior in dynamic traffic scenes by supporting both perception and decision making. In autonomous driving, most work applies world models to three areas: driving scenario generation, motion planning, and end-to-end driving.

In scenario generation, GAIA-1 \cite{hu2023gaia}, DriveDreamer \cite{wang2024drivedreamer}, and DriveWorld \cite{min2024driveworld} develop LLM-based frameworks that synthesize realistic driving content from multimodal inputs such as video, text, and actions, expanding data diversity for training. Moreover, WorldDreamer \cite{wang2024worlddreamer} treats world modeling as unsupervised visual sequence learning, emphasizing local spatiotemporal attention to strengthen dynamic understanding and speed up training convergence. Beyond scenario generation, world models also support planning and end-to-end control. Notable work such as MUVO \cite{bogdoll2023muvo} uses a world-model framework to predict future scenes based on the multimodal LiDAR and camera inputs, producing videos, point clouds, and 3D occupancy for planning. Correspondingly, MILE \cite{hu2022model} performs model-based imitation learning in CARLA, jointly learning dynamics and driving policy from offline data. SEM2 \cite{gao2022enhance} introduces a semantic-masked world model to improve sample efficiency and robustness for end-to-end control. Furthermore, Drive-WM \cite{wang2024driving} adopts a diffusion-based predictor that iteratively denoises scene representations and uses them for route planning. More recently, LAW \cite{li2024enhancing} predicts future latent scene states from current latents and ego trajectories to strengthen scene representation for end-to-end driving, while WoTE \cite{li2025end} employs a world model to extract BEV features and alleviate the scarcity of multi-future supervision. These advances demonstrate that world models can enrich data, improve trajectory prediction, and support closed-loop control. However, most approaches embed the world model within model-based pipelines without explicitly addressing rare, high-risk situations or separating perception, memory, and decision modules. In contrast, our study explores a human-inspired world-model architecture that combines the intuitive decision-making of human drivers with the contextual reasoning of large language models. To our knowledge, we are the first to apply such a world-model design to motion forecasting, using a mixture-of-expert network and LLMs to allocate capacity to corner cases while maintaining accuracy on common driving scenarios.}

{ 
\subsection{Imbalanced Learning}
Effectively learning from imbalanced data represents a significant challenge for deep learning algorithms. Highly irregular data distributions can adversely impact the model training process, leading to a substantial decline in performance when identifying and classifying minority classes. This often results in the model becoming overly aligned with the characteristics of the majority class, consequently yielding a considerable number of incorrect predictions for minority classes \cite{paoletti2023comprehensive, elreedy2024theoretical}. Such imbalances compromise the overall accuracy of the model, particularly in critical domains such as healthcare, finance, and autonomous driving, where the accurate identification of minority classes is essential.  Prior work addresses the imbalance problem at two levels \cite{chamlal2024hybrid}. Algorithm-level methods modify learning objectives or procedures, such as using cost-sensitive losses and active learning, to increase sensitivity to minority cases \cite{wang2023fendfutureenhanceddistributionaware}. Data-level methods rebalance the training set via undersampling or oversampling, sometimes with synthetic example generation \cite{upadhyay2021review}. While effective to a degree, these strategies operate largely outside the dynamics of sequential, multi-agent driving scenes. They do not explicitly model how interactions unfold over time, nor do they provide a principled way to allocate model capacity to rare interaction regimes without eroding performance on common scenarios. A complementary direction is to address the imbalance within a world model. By learning a structured latent state that explains observations, predicts scene evolution, and estimates the consequences of actions, a world model can encode scenario structure directly in its dynamics. This creates a natural locus for scenario-aware specialization. LLMs further enrich the latent state with commonsense priors and long-horizon temporal context, improving discrimination between superficially similar but behaviorally distinct scenes. Building on this insight, we employ a MoE network inside the world model’s latent space: a lightweight router directs each scene to experts specialized for particular interaction regimes, including rare corner cases, while shared components retain competence on frequent patterns. Compared with traditional algorithm- or data-level rebalancing, this design aligns specialization with the model’s learned dynamics, leverages LLM-augmented context to stabilize expert assignment, and preserves common-case accuracy by avoiding uniform reweighting that tends to overfit the tail.}

{ 
\subsection{Large Language Models for Motion Forecasting}
The field of motion forecasting for AD has evolved significantly, progressing from early physics-based models \cite{wang2025wake, wang2025dynamics} to machine learning approaches \cite{tomar2011svm}, and deep learning methods \cite{liao2024characterized, wang2025nest}. However, the complex spatiotemporal contextual information present in traffic environments, characterized by the interactions of various agents, remains a major challenge for further advancements. A promising direction emerges from the recent progress in large language models, such as Deepseek-R1~\cite{guo2025deepseek}  and Qwen \cite{bai2023qwen}. LLMs excel at multimodal reasoning, temporal abstraction, and context inference. They integrate multimodal inputs such as trajectories, maps, and agent attributes, and interpret structured traffic scenes with contextual awareness. 
Accordingly, a growing body of work incorporates LLMs into motion forecasting pipelines to enhance generalization in complex driving scenarios. For example, LLM-Traj \cite{lan2024traj} captures agent interactions without heavy prompt engineering. TR-LLM \cite{takeyama2024tr} embeds spatial reasoning to improve prediction under partial information.  Moreover, Peng et al. \cite{peng2024lc} use LLMs to enhance interpretability in lane-changing scenarios. Recent efforts \cite{liao2024gpt} also brings language-based reasoning closer to forecasting and closed-loop planning: language-informed forecasting and planning \cite{liao2024gpt}, visual–language question–answering for linking perception and control \cite{xing2025openemma}, and LLM-driven copilots and chain-of-thought approaches for interactive decision support \cite{copilot,liao2025cot}. End-to-end autonomous driving systems such as LMDrive \cite{shao2024lmdrive} process sensor streams alongside natural language instructions, and Senna \cite{jiang2024senna} decouples high-level plans from low-level trajectory generation by producing language plans that guide prediction. While these methods show that LLMs improve contextual reasoning, most treat language models as add-on modules rather than as components of a world model that explicitly represents scene state, dynamics, and the consequences of actions. As a result, they offer limited mechanisms to address safety-critical corner cases, where rare interaction regimes and occlusions demand counterfactual reasoning and strong temporal memory \cite{lan2024hi}. We take a world model view of motion forecasting and integrate LLMs into the model’s internal state. The LLM provides commonsense priors, social rules, and long-horizon context, while a lightweight temporal tokenizer maps trajectories and scene context into the LLM feature space without extra training to strengthen temporal coherence. To handle imbalance in the latent space, we pair this LLM-enhanced world model with a Mixture-of-Experts. A router directs scenes to specialized experts for distinct interaction patterns, including rare high-uncertainty cases, while shared components preserve accuracy in frequent scenarios.  In summary, we integrate LLM-based reasoning into a world model and specialize model capacity through expert routing. Instead of serving as end-to-end predictors, LLMs serve as reasoning modules, providing intent-aware context that improves adaptability in interactive scenes. This design overcomes the limited edge-case coverage and weak coupling to scene dynamics in previous LLM-based forecasting methods, resulting in a unified, robust, and context-aware AD system for safety-critical scenarios.}

\begin{figure*}
  \centering
  \includegraphics[width=0.95\linewidth]{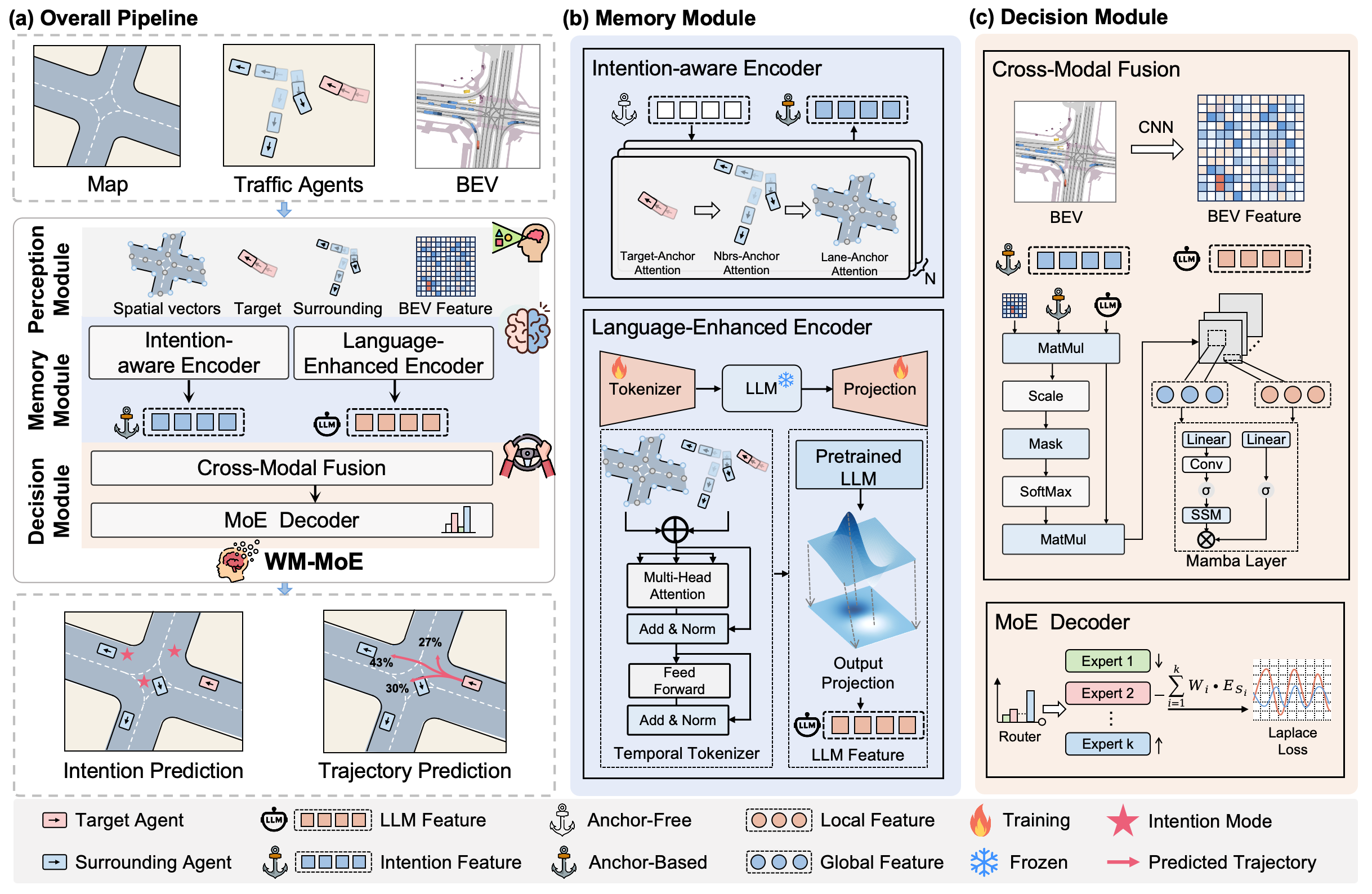}
  \caption{
  { 
Overall pipeline of WM-MoE. WM-MoE follows a world model-based pipeline with three key parts (a): Perception Module, Memory Module, and Decision Module. The Perception Module encodes traffic scenarios by capturing both spatial and temporal dynamics, resulting in a scene encoding $\bm{S}_{\text{enc}} \in \mathbb{R}^{T \times B \times D_{\text{emb}}}$. This encoding is fed into the Memory Module (b), which includes the Intention-Aware Encoder and the Language-Enhanced Encoder, producing multimodal queries $\bm{q}_{\text{mode}}$ and language-informed features $\bm{t}_{\text{llm}}$, respectively. In the Decision Module (c), a Cross-Modal Fusion employs cross-attention to integrate $\bm{q}_{\text{mode}}$, $\bm{t}_{\text{llm}}$, and the BEV feature $\bm{v}_{\text{enc}}$ (extracted via a CNN backbone), yielding the enhanced cross-modal feature $\bm{f}_{\text{c}}$. Finally, the MoE Decoder utilizes the cross-modal feature to generate the candidate multimodal future trajectories.}
}
  \label{fig1}
\end{figure*}

\section{METHODOLOGY}\label{Method}

\subsection{Problem Formulation}
The goal of this study is to accurately forecast the future path of a target agent within the perception range of an AV. Consider a traffic scenario involving \(n\) traffic agents (e.g., vehicles, pedestrians, and bicycles) surrounding the AV. Let \(x_i^{(-t_h-1):0}\) represent the historical state of the \(i\)th agent (\(i \in [0,n]\)), encompassing its 2D position coordinates, velocity, acceleration, and yaw angle over a specified time horizon \(t_h\). Given the historical states of the target agent (indexed as 0) and its \(n\) surrounding agents (indexed from 1 to \(n\)), collectively represented as \(X = \{x_0^{(-t_h-1):0}, x_1^{(-t_h-1):0}, ..., x_n^{(-t_h-1):0}\}\), along with the corresponding high-definition (HD) map $M$ and bird's-eye view (BEV) information $I$ of the traffic scene, the model is meticulously designed to predict the future state of the target agent over the next \(t_f\) time steps, denoted as \(Y = \{y^{1:t_f}\}\).

{ 
\subsection{Model Architecture}
As shown in Figure \ref{fig1}, WM-MoE adopts a world model architecture inspired by human cognition with three modules: Perception, Memory, and Decision. Initially, the Perception Module encodes traffic scenarios from input data into structured scene encodings that capture spatial and temporal dependencies. These encodings are then processed by the Memory Module, which leverages a frozen LLM and employs attention to aggregate interactions and map semantics into language-informed features. In the Decision Module, a cross-modal fusion module integrates the encoded contextual features to generate cross-modal representations. Next, an MoE router then operates in the latent space, computing fine-grained similarities between representations and scoring tokens, and dispatching them to specialized expert networks accordingly. Finally, the MoE Decoder combines the top-$k$ expert outputs to produce multimodal future trajectories conditioned on the combined scene and cross-modal features.}

\subsection{Perception Module}
{ 
The Perception Module serves as the entry point of the world model, constructing the latent scene state used for prediction and counterfactual rollouts. At each time step, it compresses multimodal inputs into compact, agent- and map-aware tokens that preserve spatial layout, temporal evolution, and interaction cues for downstream Memory and Decision Modules, yielding an abstract yet behaviorally sufficient representation of the physical world’s key concepts and relations. In line with established practices for sequence-to-sequence networks, the Perception Module utilizes a combination of Multi-Layer Perceptrons (MLPs), Gated Recurrent Units (GRUs), and attention mechanisms to hierarchically embed the input data. Specifically, given the historical states of the target agent \(x_0^{(-t_h-1):0}\) and the surrounding agents \(x_{1:n}^{(-t_h-1):0}\) within a traffic scene, we utilize a stack of parameter-unshared GRUs and MLPs to hierarchically embed the position vectors of traffic agents and extract the corresponding temporal vectors \(\bm{t}_{\text{enc}} \in \mathbb{R}^{B \times T \times D_{\text{emb}}}\) for the target agent and its surrounding agents \(\bm{n}_{\text{enc}} \in \mathbb{R}^{B \times (N_{veh}+ N_{ped})\times D_{\text{emb}}}\). Here, \(D_{\text{emb}}\) is the dimension of the hidden layer, while \(B\) denotes the batch size, and \(T\) represents the sequence length corresponding to the past \(t_h\) seconds. \(N_{veh}\) and \(N_{ped}\) signify the number of vehicles and pedestrians, respectively. In parallel, for the corresponding HD maps \(M\), we adopt the same backbone proposed in HiVT \cite{zhou2022hivt} to represent lane nodes and lane lines as discrete vectors. These vectors are fed into MLPs to capture spatial vectors \(\bm{l}_{\text{enc}} \in \mathbb{R}^{N_l \times B \times D_{\text{emb}}}\), where \(N_l\) represents the maximum number of lane nodes. Accordingly, a CNN backbone extracts BEV features, BEV feature $\bm{v}_{\text{enc}}$ to capture local geometry, and drivable context complementary to the vectorized map. Furthermore, these spatial representations are independently fed into separate self-attention mechanisms to model dense connections, further extracting and continuously updating the features. This process results in an embedded representation with spatio-temporal awareness, ultimately aggregated into a scene representation \(\bm{S}_{\text{enc}} \in \mathbb{R}^{T \times B \times D_{\text{emb}}}\).}

\subsection{Memory Module}
{ The Memory Module maintains and updates the latent scene state of the world model. It accumulates evidence over time, integrates multi-agent interactions, and produces future-oriented encodings that support long-term prediction and counterfactual rollouts. Concretely, it reads perceptual tokens and writes a temporally consistent state \( \bm{S}_{\text{enc}} \) by combining intention-aware querying with language-informed temporal reasoning.}

\subsubsection{Intention-Aware Encoder}

This module is responsible for extending the unimodal scene representation \( \bm{S}_{\text{enc}} \) into a multimodal form. Inspired by the DETR framework \cite{carion2020endtoend}, we adopt a similar approach, employing learnable multimodal queries as additional inputs to the module. We incorporate attention mechanisms and positional encoding to progressively refine and capture the spatiotemporal interactions between the agents and environmental elements.

Technically, we begin by defining anchor queries \(\bm{q} \in \mathbb{R}^{K_n \times T \times B \times D}\) and reshape the scene representation \( \bm{S}_{\text{enc}} \) into \(K_n\) modalities. Positional encoding is then applied to map the query states to a high-dimensional space, embedding spatial positional information, which results in \(\bm{t}_{\text{rep}} \in \mathbb{R}^{K_n \times T \times B \times D}\). Subsequently, we introduce a novel cross-attention mechanism designed to iteratively enhance and update the representation of the query states. Specifically, the cross-attention mechanism \(\mathcal{A}(\cdot)\) is designed to uncover the relationships between each pair of elements in the sequence—the query state \(\bm{q}\), the embedded scene representation \(\bm{t}_{\text{rep}}\), spatial vectors \(\bm{l}_{\text{enc}}\), and the context information of surrounding agents \(\bm{n}_{\text{enc}}\). Following this, the updated query state and cross-attention output are integrated and normalized through layer normalization \(\mathcal{N}(\cdot)\) to mitigate the risk of forgetting past information and enable the network to learn long-term dependencies of historical trajectories.
Formally, 
\begin{equation}
    \bm{q}_{\text{scene}} = \bm{q} + \mathcal{N}(\mathcal{A}(\bm{q}, \bm{t}_{\text{rep}} + \text{PE}(\bm{q}), \bm{t}_{\text{rep}}))
\end{equation}
\begin{equation}
    \bm{q}_{\text{spatial}} = \bm{q}_{\text{scene}} +  \mathcal{N}(\mathcal{A}(\bm{q}_{\text{scene}}, \bm{l}_{\text{enc}} + \text{PE}(\bm{q}_{\text{scene}}), \bm{l}_{\text{enc}}))
\end{equation}
\begin{equation}
    \bm{q}_{\text{mode}} = \bm{q}_{\text{spatial}} + \mathcal{N}(\mathcal{A}(\bm{q}_{\text{spatial}}, \bm{n}_{\text{enc}} + \text{PE}(\bm{q}_{\text{spatial}}), \bm{n}_{\text{enc}}))
\end{equation}
where $\text{PE}(\cdot)$ denotes the position encoding mehancism.
Furthermore, the multimodal query \(\bm{q}_{\text{mode}}\) is reshaped and expanded to \(\bm{q}_{\text{mode}} \in \mathbb{R}^{t_f \times K_n \times B \times D_{\text{emb}}}\), aligning it with the temporal steps of the future trajectory distribution.

\subsubsection{Language-Enhanced Encoder}{ 
{ 
In real-world driving, autonomous vehicles must interpret traffic rules and social norms, infer the latent intentions of surrounding agents under ambiguous or sparse observations, and maintain temporal consistency across long horizons—all while contending with long-tail data imbalance and distribution shift. These challenges are particularly acute in corner-case scenarios, where accurate prediction depends on reasoning over right-of-way constraints, map semantics, and subtle agent interactions. However, conventional supervised models primarily learn from frequent patterns and often lack explicit mechanisms to incorporate rule-based or intent-aware priors. Similarly, imitation learning pipelines, though effective in common cases, are prone to compounding errors and policy drift in previously unseen interactions. To address these limitations, we introduce the Language-Enhanced Encoder, designed to improve the model’s scene understanding and reasoning capacity.

Within the world model, this module serves as the temporal reasoning and priors component. Its purpose is to update the latent scene state by combining sequential evidence with commonsense constraints, social rules, and intent cues that are difficult to learn from skewed driving data. In practice, the module aligns trajectories and context with a pretrained LLM so that the state transition of the world model is informed by language-derived priors. This improves long-horizon consistency, supports counterfactual rollouts, and strengthens discrimination in ambiguous scenes.

At its core is a temporal tokenizer, which projects sequential motion data into the latent space of a pretrained LLM, thereby enabling alignment between temporal dynamics and the rich, contextual priors embedded in large-scale language corpora. Methodologically, this tokenizer produces encodings for both the target agent \(\bm{t}_{\text{enc}}\) and its surrounding agents \(\bm{n}_{\text{enc}}\). To mitigate data offset and numerical instability, we apply normalization on these encodings and obtain normalized states $\bm{t}_{\text{enc}}^{\text{norm}}$ and $\bm{n}_{\text{enc}}^{\text{norm}}$. The temporal tokenizer prepares comprehensive spatiotemporal language-informed features $\bm{t}_{\text{llm}}$ for the LLM by concatenating the normalized states and feeding them into MLP layers. After processing by the LLM, the final component of the module is an output projection layer. This projection employs a two-layer MLP to align the dimensionality of the LLM-generated features with that of the original modality, ensuring compatibility with subsequent modules. The aligned data, along with the processed BEV information and query data, is then passed to the cross-modal fusion component. Furthermore, the resulting projection maps the LLM features back to the world model’s latent dimensionality, ensuring compatibility with cross-modal fusion and subsequent decision dynamics.

Notably, this study refrains from employing large-scale models such as LLaMMA-3 \cite{touvron2023llama} or GPT-4 to address the stringent real-time requirements of AD systems. Although large LLMs often deliver strong cross-domain performance, their role in motion forecasting is primarily to provide semantic context and to assist latent world-state representation rather than to act as end-to-end predictors.  Furthermore, these LLMs are primarily designed for natural language processing and computer vision tasks, necessitating extensive training or fine-tuning on massive datasets to perform motion prediction effectively. However, in the domain of motion prediction—particularly under edge cases—the availability of high-quality training data is inherently limited. This constraint renders such models less suitable for highly specialized and safety-critical applications.

These constraints make very large LLMs unsuitable for on-vehicle deployment where compute, latency, and energy budgets are tight. Instead, we deliberately avoid very large LLMs and opt for a lightweight GPT-2 backbone coupled with the temporal tokenizer. This choice aligns with the requirements of a driving world model: state updates must run in real time, data for rare events is limited, and robustness under distribution shift is essential. The design injects language-derived structure and semantic priors into the latent state without extensive task-specific retraining, keeps computation bounded, and preserves stability. In practice, the lightweight LLM provides sufficient semantic context, and the tokenizer aligns trajectories and scene cues with temporal dynamics, achieving a favorable balance between accuracy and efficiency. We provide a detailed quantitative comparison of LLM backbones and the contribution of the temporal tokenizer in the Experiments section.

}

}

\subsection{Decision Module}
{ In the world model, the Decision Module operates on the latent scene state to produce action-conditioned futures and to support counterfactual rollouts. Its role is to read the memory state, fuse complementary modalities, and simulate how the scene is likely to evolve under different interaction regimes. In a nutshell, the module first forms a cross-modal latent that is aligned with the world model state, then applies expert specialization to generate trajectory hypotheses that are consistent with map semantics, social rules, and agent dynamics.}

\subsubsection{Cross-modal Fusion}
This component is engineered to process and integrate the multimodal query \(\bm{q}_{\text{mode}}\), the LLM-enhanced vector \(\bm{t}_{\text{llm}}\), and the BEV visual information \(\bm{v}_{\text{enc}}\), to capture cross-modal interactions effectively. Specifically, for the BEV image, we utilize a CNN backbone to extract visual features \(\bm{v}_{\text{enc}} \in \mathbb{R}^{C \times H \times W}\), where \(C\), \(H\), and \(W\) represent the channel, height, and width of the BEV image, respectively. To enhance the model's robustness against environmental perturbations, we introduce independent noise terms \( z_q \sim \mathcal{N}(0, 1) \), \( z_l \sim \mathcal{N}(0, 1) \), and \( z_v \sim \mathcal{N}(0, 1) \), which are applied through fully connected layers to the input vectors \(\bm{q}_{\text{mode}}\), \(\bm{t}_{\text{llm}}\), and \(\bm{v}_{\text{enc}}\), respectively. This process produces perturbed outputs \(\bm{\bar{q}}_{\text{mode}}\), \(\bm{\bar{t}}_{\text{llm}}\), and \(\bm{\bar{v}}_{\text{enc}}\). These perturbed vectors are then fused using the attention mechanism, which allows the multimodal query \(\bm{\bar{q}}_{\text{mode}}\) to integrate knowledge from both the LLM-enhanced vector \(\bm{\bar{t}}_{\text{llm}}\) and the BEV visual features \(\bm{\bar{v}}_{\text{enc}}\). The result of this fusion is the cross-modal feature \(\bm{q}_{\text{c}}\).  Formally,
\begin{equation}
\bm{q}_{\text{c}} = \mathcal{A}( \bm{\bar{t}}_{\text{llm}}, \bm{\bar{v}}_{\text{enc}}, \bm{\bar{q}}_{\text{mode}})
\end{equation}

Following this, the feature \(\bm{q}_{\text{c}}\) is processed through a combination of 1D and 2D Temporal Convolutional Networks (TCNs) to extract temporal dependencies. Formally, the intermediate representation is updated as follows:
\begin{equation}
\bm{q}_{\text{c}}' = \sigma (\phi_{\textit{TCN1D}} (\bm{q}_{\text{c}})) + \phi_{\textit{TCN2D}} (\bm{q}_{\text{c}})
\end{equation}
where \(\phi_{\textit{TCN1D}}\) and \(\phi_{\textit{TCN2D}}\) represent the transformations performed by the 1D and 2D TCNs, respectively, and \(\sigma\) denotes the LeakyReLU activation function.

{ 
After attention-based fusion yields $\bm{q}_{\mathrm{c}}$ and 1D/2D TCNs aggregate local temporal and spatial patterns into $\bm{q}_{\mathrm{c}}'$, we then insert a lightweight, linear-time Mamba framework \cite{dao2024transformers} refinement in place of stacking deeper TCNs (fixed receptive fields) or extra Transformers (quadratic complexity) to prepare this signal for expert routing. Thanks to selective state-space dynamics, Mamba captures long-range, non-stationary dependencies and suppresses cross-modal conflicts that arise under missing/occluded inputs, thereby stabilizing the representation passed to the MoE. Concretely, we refine $\bm{q}_{\mathrm{c}}'$ with two Mamba branches followed by a parameter-free pointwise projection $\phi_{\textit{MLP}}$, and aggregate them:
\begin{equation}
\bm{f}_{\mathrm{c}}
= \phi_{\textit{MLP}}\!\big(\mathcal{M}_{1}(\bm{q}_{\mathrm{c}}')\big)
+ \phi_{\textit{MLP}}\!\big(\mathcal{M}_{2}(\bm{q}_{\mathrm{c}}')\big),
\end{equation}
where $\mathcal{M}_{1}$ and $\mathcal{M}_{2}$ are Mamba networks instantiated with $3{\times}3$ and $1{\times}1$ convolutional kernels, respectively, to jointly model context-sensitive propagation (3{\texttimes}3) and channel-wise recalibration (1{\texttimes}1). Next, the resulting enhanced feature $\bm{f}_{\mathrm{c}}$ serves as the comprehensive input to the MoE Decoder for motion forecasting.}

{ 
 
\subsubsection{MoE Decoder}

\begin{figure}
    \centering
    \includegraphics[width=0.7\linewidth]{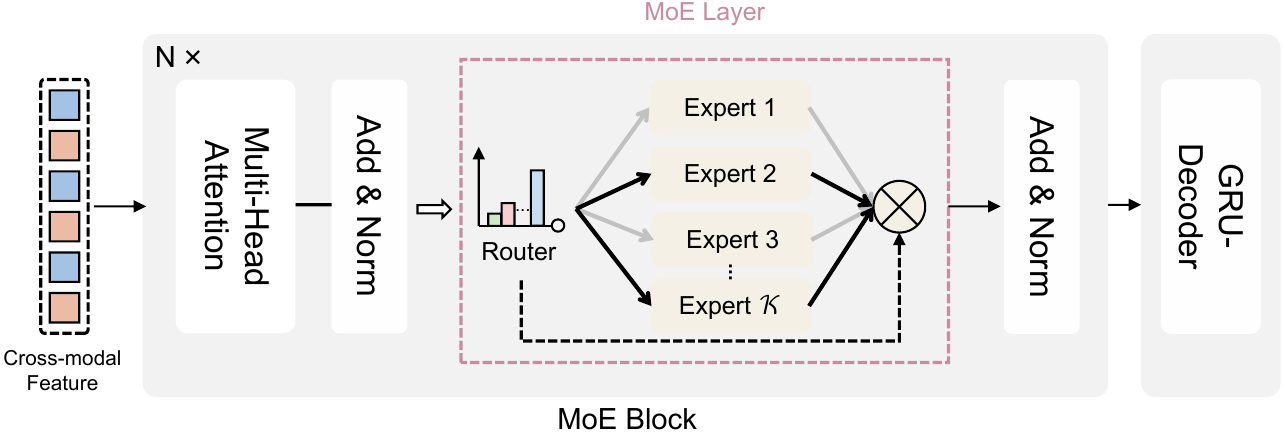}
    \caption{Overall architecture of the MoE decoder. It comprises (\(N\)) stacked MoE blocks, each with a gating layer that routes inputs to ($K$) experts. The decoder aggregates contributions from multiple experts rather than relying on a single expert for any scenario.}
    \label{fig:MoE Decoder}
\end{figure}

The MoE Decoder is the simulation head of the world model. It generates future trajectories by routing tokens in the latent space to specialized experts that model distinct interaction patterns, including corner cases.  In our approach, we implement a MoE-based Transformer architecture, where the proposed MoE layer is incorporated into each layer of the Transformer decoding block, replacing the standard Feed-Forward Network (FFN). As shown in Figure \ref{fig:MoE Decoder}, the proposed MoE layer aggregates the outputs of all experts, with the routing mechanism dynamically assigning scores to weight their contributions. Unlike traditional MoE implementations that activate only a subset of experts, this design ensures that the decoder fully exploits the diverse knowledge embedded in all the experts. The detailed calculation process of the $N$ block configured with the MoE layer is as follows:
\begin{equation}
\begin{aligned}
        \bm{h}_l^s &= \phi_{\textit{MSA}}(\phi_{\textit{LN}}(\bm{h}_{l-1})) + \bm{h}_{l-1} \quad l = 1, \dots N\\
        \bm{h}_l^m &= \phi_{\textit{MoE}}(\phi_{\textit{LN}}(\bm{h}_l^s)) + \bm{h}_l^s \quad l = 1, \dots N\\
        \bm{h}_l &= \phi_{\textit{LN}}(\bm{h}_l^m) \quad l = 1, \dots N \\
\end{aligned}
\end{equation}
where \(\phi_{\textit{MSA}}\) and \(\phi_{\textit{LN}}\) denote the multi-head self-attention and layer normalization, respectively. \(h_l\) represents the output of the \(l\)-th MoE block, with the initial input defined as \(\bm{h}_0 = \bm{f}_c\). \(h_l^s\) and \(h_l^m\) represent the outputs of the self-attention and MoE layers, respectively, in the \(l\)-th MoE block. Within the MoE layer, the router is implemented as a linear function that predicts the probability of assigning each token to a specific expert. The outputs of the selected experts are aggregated based on their respective gating weights. The overall computation process can be expressed as follows:  
\begin{equation}
    \bm{p}_i = \frac{\exp(\bm{W}_i \cdot \bm{h}_l^s)}{\sum_{j=1}^{\mathcal{K}} \exp(\bm{W}_j \cdot \bm{h}_l^s)} \quad  i = 1, 2, \ldots, \mathcal{K}
\end{equation}
where \(\bm{p}_i\) is the selected logits to distribute selection weights among the experts, and  
 \(\mathcal{K}\) is the total number of experts in each MoE layer. The contributions of the selected experts are then aggregated and weighted by the normalized logits:
\begin{equation}
    \bm{h}_l^m = \sum_{i=1}^{\mathcal{K}} \bm{p}_i \cdot \textit{E}_i(\phi_\textit{LN} (\bm{h}_l^s)) + \bm{h}_l^s \quad l = 1, \dots N
\end{equation}
where each expert network is denoted by \( \textit{E}_i \). After passing through \(N\) MoE blocks, the resulting feature \(\bm{h}_N = \bm{f}_{\text{moe}}\) is fed into a GRU layer to capture temporal dependencies in the data. The decoder processes both the aggregated features and target information to generate output feature representations. The decoder ultimately predicts the future trajectory $\hat{y}^{1:t_f}$ characterized by a Laplace distribution. This approach serves to enhance the robustness of the model, enabling the decoder to adaptively leverage the input of specialized experts, thus allowing it to respond effectively to the specific characteristics of distinct scenarios.

Overall, our rationale for designing the MoE decoder to specialize based on scenarios rather than static elements like road features is rooted in the fundamental nature of the motion forecasting challenge. The most difficult prediction problems arise not from the road geometry itself, but from the complex, dynamic interactions and behaviors of agents within that environment. A corner case is defined by an event—such as a sudden lane change, an abrupt stop, or an intricate maneuver at an intersection—not by the static background. Therefore, we employ a ``divide and conquer'' strategy where each expert in the MoE network learns to model the nuanced dynamics of a specific class of scenarios. This allows the model to dedicate specialized capacity to handling rare but critical events, without degrading performance on more common driving patterns. This design breaks down the complex, multimodal space of future trajectories into simpler, more manageable sub-problems. Each sub-problem is then handled collaboratively by multiple dedicated specialists.
}

{ 
\subsection{Training}
To ensure the diversity and accuracy of motion forecastings, we employ a multitask learning strategy that incorporates both motion forecasting losses and maneuver classification losses, tailored specifically to the characteristics and evaluation metrics of each dataset. 

For the nuScenes dataset, we adopt the Laplace Negative Log-Likelihood as the primary regression loss \(\textit{L}_{reg}\), along with Cross Entropy loss \(\textit{L}_{cls}\) for mode classification. These losses are combined with the motion forecasting loss \( \textit{L}_{ade} \), which evaluates the Average Displacement Error (ADE) over the predicted trajectories. Formally,
\begin{equation}
    \textit{Loss} = \textit{L}_{ade} + \lambda_1 \cdot \textit{L}_{reg} + \lambda_2 \cdot \textit{L}_{cls},
\end{equation}
where \(\textit{L}_{ade}\) represents the ADE metric, while \(\lambda_1\) and \(\lambda_2\) are weighting coefficients used to balance the contributions of regression and classification objectives. The Laplace Negative Log-Likelihood ensures that the predicted trajectories match the ground truth distribution, while the Cross-Entropy loss supervises the model's ability to classify different maneuver modes, enabling robust multimodal prediction.

For the NGSIM, HighD, and MoCAD datasets, where Root Mean Square Error (RMSE)  is the standard evaluation metric, we adopt the decoder and loss calculation strategies described in the classic work \cite{liao2024bat}. Specifically, the Mean Squared Error (MSE) loss (\(\textit{L}_{mse}\)) is applied as the motion forecasting loss and combined with the Cross Entropy loss (\(\textit{L}_{ce}\)) for maneuver classification. The final loss function for these datasets is defined as follows:
\begin{equation}
    \textit{Loss} = \gamma_1 \cdot \textit{L}_{mse} + \gamma_2 \cdot \textit{L}_{ce},
\end{equation}
where \(\gamma_1\) and \(\gamma_2\) are weighting coefficients. The use of MSE loss ensures alignment with dataset-specific evaluation metrics, while Cross Entropy loss enhances the model's multimodal reasoning capabilities.  Notably, the selection of loss functions primarily follows the evaluation metrics defined in the official protocols of each dataset. This ensures alignment with the standard benchmarks and maintains consistency with prior work using these datasets, thereby guaranteeing fairness in performance comparisons.}

\section{Experiment}\label{Experiment}

\subsection{Experimental Setups}
To thoroughly demonstrate the capabilities of our proposed WM-MoE model, we conduct a series of experiments, each designed to target a specific task. These experiments include (1) accuracy evaluation on four real-world driving datasets; (2) robustness and adaptability assessment in corner-case and data-missing scenarios; (3) comprehensive comparisons against other methods; and (4) ablation studies on real data to evaluate the contributions of individual components. Furthermore,  we assess the impact of different LLMs (5) and varying the number of expert networks in the MoE module (6) to verify the effectiveness of our design; and (7) finally, we present qualitative results and failure cases under rare complex traffic conditions to visually compare the model's prediction accuracy and adaptability, and indicate the limitations and avenues for future improvement.

{ 
\subsection{Datasets}

This study conducts extensive experiments using several real-world driving benchmarks, including nuScenes \cite{caesar2020nuscenes}, NGSIM \cite{8575356}, HighD \cite{8569552}, and MoCAD \cite{liao2024bat} datasets. All datasets follow standard segmentation protocols from prior work or official releases. Furthermore, we evaluate our model on the newly proposed nuScenes-corner dataset to assess its performance in corner-case scenarios and validate its robustness in data-missing scenes.

Importantly, this study presents a new dataset, called ``nuScenes-corner'', specifically designed to evaluate the performance of motion forecasting models in rare and unbalanced scenarios. This dataset is derived from the nuScenes dataset, where we analyze the TTC metric of the target vehicle across all scenes. Specifically, for the subset of high-risk scenarios (TTC $<$ 2s), we select 1,120 turning maneuvers (including U-turns), 984 congested environments, 456 sudden acceleration scenarios, and 832 abrupt braking instances. These represent corner-case scenarios that are essential for evaluating the model's performance under safety-critical conditions.  For turning scenarios, we identify scenarios based on the yaw angle of the target vehicle. A scenario is classified as a left or right turn when the absolute value of the yaw angle exceeds 0.3 radians, whereas a U-turn is identified when the yaw angle surpasses 0.7 radians. This precise classification allows for detailed evaluation of a model's capacity to handle substantial directional changes, which are crucial for safe navigation in urban environments with intersections and complex turns. Moreover, congested scenarios are defined by the density of nearby vehicles and pedestrians: when there are more than 35 vehicles or over 50 pedestrians, the scenario is considered congested. Evaluating the model under these conditions provides insight into its capability to navigate dense traffic, common in urban areas during peak hours or near busy intersections. In addition, for sudden acceleration and abrupt braking scenarios, classification depends on notable changes in speed and acceleration. Abrupt braking is characterized by a rapid speed reduction with negative acceleration exceeding a threshold, while sudden acceleration involves a rapid speed increase with positive acceleration beyond the set limit. These scenarios assess the model's responsiveness to sudden driving changes, which are prevalent in real-world conditions due to unexpected obstacles or shifting traffic patterns. Table \ref{case_study_2} showcases a statistical summary of these categories, elucidating their distinct kinematic signatures. For instance, the Turning scenarios exhibit a high average yaw rate (0.6765 rad/s) at low speeds, while Braking scenarios are defined by a strong negative average acceleration of -1.96 m/s². Conversely, Acceleration events show a high positive acceleration of 3.57 m/s² and the highest average speed, and Congested scenes are characterized by very low speeds (8.58 km/h) and minimal acceleration, typical of stop-and-go traffic. Furthermore, Figure \ref{fig:distri} visualizes the joint spatial distributions of these events. The plots reveal that corner cases are not randomly distributed across the map but are highly concentrated in specific regions. For example, turning and braking events show dense clusters that likely correspond to intersections, while acceleration patterns might correlate with intersection exits or on-ramps. This spatial concentration underscores the necessity for models to possess a strong understanding of scene context and road topology to anticipate these challenging behaviors. 
Overall, this dataset provides a comprehensive benchmark to test the model's ability to manage complex, real-world driving situations effectively, offering valuable insights into their practical applicability for AVs.

\begin{table}[ht]
  \centering
  \caption{ {Statistical Analysis of the nuScenes-Corner dataset. The table summarizes the key properties for each of the four identified corner-case scenarios. The distinct values for average speed, acceleration, and yaw rate highlight the unique kinematic signatures of each event type, validating their classification as challenging, non-nominal driving behaviors. AVG. Speed: average speed; Avg. Acceleration: average acceleration; Avg. Yaw Rate: average yaw rate.}}
      \setlength{\tabcolsep}{1.5mm}
   \resizebox{0.75\linewidth}{!}{
    \begin{tabular}{c|cccc}
 \bottomrule 
    Scenario & Sample Size &AVG. Speed (km/h) & Avg. Acceleration (m/$s^2$) & AVG. Yaw Rate (rad/s) \\
  \hline
    Turning   & 1120 &  12.5981 & 0.4327 & 0.6765\\
    Congested & 984 & 8.5784 & 0.1627 & 0.0378\\
    Braking & 832 & 10.3724 & -1.9584 & 0.0724\\
    Acceleration & 456 & 34.7651 & 3.5724 & 0.0583\\
 \toprule
    \end{tabular}
  \label{case_study_2}
  }
\end{table}

\begin{figure}
\centering
\includegraphics[width=\textwidth]{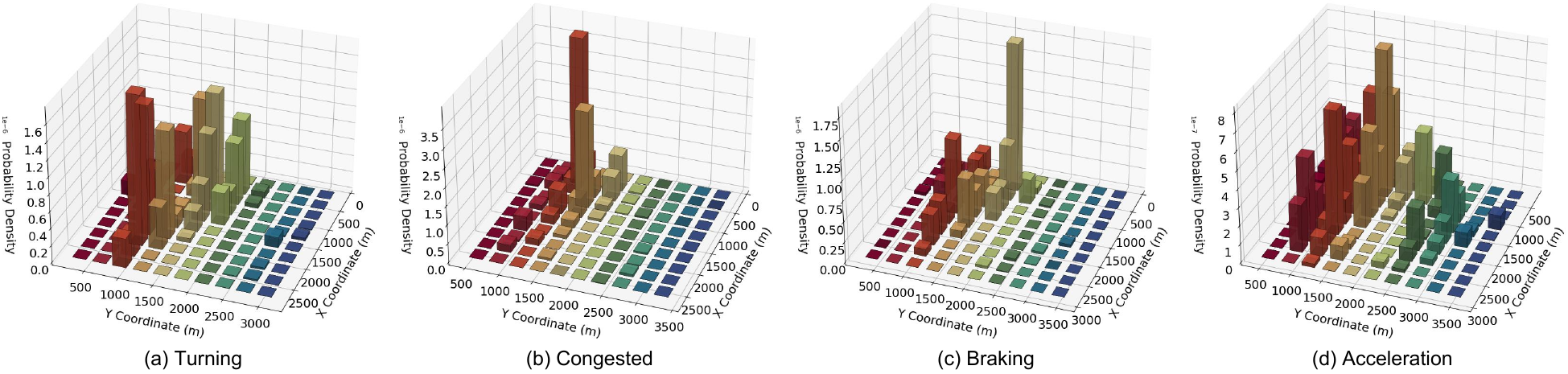}
\caption{
{ {Joint spatial distributions of corner-case scenes in the nuScenes-corner dataset. The plots visualize the probability density (z-axis) of (a) Turning, (b) Congested, (c) Braking, and (d) Acceleration events over the map's x-y coordinates. The high-density clusters (z-axis) reveal that these critical events are not uniformly distributed but are highly concentrated in specific geographic areas, suggesting a strong correlation between road topology and the occurrence of corner cases.}}}
\label{fig:distri}
\end{figure}

To comprehensively evaluate our model's robustness and its ability to handle complex driving dynamics and scene interactions, we consider the issue of missing data. Specifically, we design specialized variants of the nuScenes test set to simulate different levels of data loss and evaluate our proposed model's robustness. Based on the original historical trajectory data, we construct missing data scenarios with 20\%, 40\%, and 60\% frame loss, corresponding to one-frame, two-frame, and three-frame drop rates. For each scenario, we apply a randomized mechanism to determine the missing observations, ensuring unbiased data removal and maintaining consistent input dimensionality. This method better reflects the inherent unpredictability of real-world data collection. Missing frames are replaced with zero values to represent the absence of historical state observations. These variants are denoted as nuScenes (drop 1 frame), nuScenes (drop 2 frames), and nuScenes (drop 3 frames). This approach simulates real-world conditions, where sensors may fail to perceive surrounding agents due to occlusions, adverse weather, or other environmental constraints.

Notably, NGSIM and HighD provide local vehicle and map coordinates, while nuScenes, nuScenes-corner, and MoCAD also include high-definition semantic maps, satisfying the spatial reasoning needs of our model. We apply these datasets as benchmarks to evaluate our model's accuracy and robustness under various traffic conditions.}

\subsection{Evaluation Metrics}
{ 
The model’s performance is assessed using different metrics tailored to the datasets. For nuScenes dataset, we use the Minimum Average Displacement Error over $G$ trajectories (\(\text{minADE}_g\)), the Miss Rate within 2 meters for the top K trajectories (\(\text{MR}_{g}\)), and the Final Displacement Error (\(\text{minFDE}_g\)) over $G$ trajectories, which can be defined as follows:
\begin{equation}
\operatorname{minADE}_g=\frac{1}{N} \sum_{i=1}^N \min_{g=1}^G\left(\frac{1}{t{_f}} \sum_{t=1}^{t_f}\left\|\hat{y}_{i, g}^t-y_i^t\right\|_2\right)
\end{equation}
\begin{equation}
\operatorname{minFDE}_g=\frac{1}{N} \sum_{i=1}^N \min _{g=1}^G\left\|\hat{y}_{i, g}^t-y_i^t\right\|_2
\end{equation}
\begin{equation}
\operatorname{MR}_g=\frac{1}{N} \sum_{i=1}^N \textbf{1}(\min _{g=1}^G\left\|\hat{y}_{i, g}^t-y_i^t\right\|_2 > \theta_M)
\end{equation}
where $N$ denotes the number of samples, $\mu_{i, g}^t $ is the predicted position of the $g$-th intention mode in the $i$-th sample at time step $t$, and $x{_0}_i^t$ is the ground truth position. $\|\cdot\|_2$ denotes the Euclidean norm. \textbf{1} represents the indicator function. For the nuScenes dataset, the threshold $\theta_M$ is set to 2.0 meters. Moreover, we utilize Root Mean Squared Error ($\operatorname{RMSE}_g$) as the evaluation metric for the NGSIM, HighD, and MoCAD datasets. Mathematically,
\begin{equation}
\operatorname{RMSE}_g=\sqrt{\frac{1}{N} \sum_{i=1}^N \min _{g= 1}^G\left(\frac{1}{t_f} \sum_{t=1}^{t_f}\left\|\hat{y}_{i, g}^t-y_i^t\right\|_2\right)}
\end{equation}}

\subsection{Training and Implementation Details}
WM-MoE is trained on an NVIDIA RTX 4090 GPU with 24 GB memory. We use the Adam optimizer with a learning rate of \(5 \times 10^{-4}\) and the CosineAnnealingLR scheduler (\(T_{\text{max}} = 150\), \(\eta_{\text{min}} = 5 \times 10^{-6}\)). The loss function parameters \(\lambda_1\) and \(\lambda_2\) are set to 1 and 0.5, respectively. 
For the NGSIM, HighD, and MoCAD datasets, we follow the STDAN encoder \cite{9767719} with a hidden layer dimension of 64, using \(\gamma_1 = \gamma_2 = 1\). Training configurations vary across datasets. For nuScenes, we use a batch size of 32 and train for 120 epochs. For NGSIM, HighD, and MoCAD, we adopt a batch size of 128 and train for 12 epochs. The key implementation details and model parameters are as follows:

\textbf{Perception Module.}
This component applies an MLP, Transformer encoder, and GRU to encode the target vehicle's input data into a hidden state \(\bm{t}_{\text{enc}} \in \mathbb{R}^{B \times T \times D_{\text{emb}}}\), with \(D_{\text{emb}} = 64\). A similar approach encodes surrounding vehicles and pedestrians using a dynamic GRU to handle agents present at different time frames, forming \(\bm{n}_{\text{enc}} \in \mathbb{R}^{B \times (N_{\text{veh}} + N_{\text{ped}}) \times D_{\text{emb}}}\). Lane geometry is discretized into segments and encoded via an MLP followed by self-attention, resulting in \(\bm{l}_{\text{enc}} \in \mathbb{R}^{N_l \times B \times D_{\text{emb}}}\). Pairwise attention is then applied between the target vehicle, lane segments, and surrounding agents, yielding the final scene encoding \(\bm{S}_{\text{enc}} \in \mathbb{R}^{T \times B \times D_{\text{emb}}}\). Attention layers use 4 heads with a 0.1 dropout rate, while lane lines are encoded using 2 GAT layers.

\textbf{Memory Module.}
For the Intention-Aware Encoder, we initialize anchor queries \(\bm{q} \in \mathbb{R}^{K_n \times B \times D_{\text{emb}}}\), where \(K_n\) represents the number of predicted modes, \(B\) is the batch size, and \(D_{\text{emb}} = 64\) is the hidden dimension. We initialize positional encoding with the same hidden size and a scale factor of 10,000 to embed positional information. The subsequent attention mechanism uses 4 attention heads, each with a hidden size of 64. Moreover, we use a two-layer MLP architecture in both the tokenizer and output projection for the Language-Enhanced Encoder. The hidden layer dimensions are set to 64, with a dropout rate of 0.1. The output dimensions vary according to the LLM used—GPT-2, GPT-Neo, and TinyLlama—with MLP output dimensions of 1024, 2048, and 4096, respectively. 

\textbf{Decision Module.}
In the Cross-modal Fusion, we integrate the multimodal query \(\bm{q}_{\text{mode}}\), the LLM-enhanced vector \(\bm{t}_{\text{llm}}\), and BEV information \(\bm{v}_{\text{enc}}\). BEV data is processed using two 2D convolutional layers with a kernel size of 4, followed by three additional 2D convolutional layers with a kernel size of 3, and all layers use a dropout rate of 0.1, resulting in a BEV vector of size \(3 \times 750 \times 750\). We then employ an attention layer with 4 heads and a hidden dimension of 64 to fuse information from these modalities. Next, we apply 1D and 2D TCNs to capture different-level features. The 1D TCN utilizes multiple convolutional layers, with dimensions matching a hidden size of 64, and incrementally increases dilation factors to expand the receptive field, enabling effective modeling of temporal dependencies. The 2D TCN processes the global feature map, employing convolution along spatial dimensions to extract higher-level spatio-temporal characteristics. Both TCN types maintain input and output dimensions at the hidden size, and their dilation factors progressively increase to capture long-range dependencies.
To achieve an optimal balance between depth and computational cost, we have incorporated skip connections after each convolutional operation. This approach helps mitigate the issue of vanishing gradients and accelerates training, thereby enhancing the model's robustness and efficiency in capturing complex temporal-spatial interactions.
Additionally, the MoE Decoder employs an MoE layer to handle complex scenarios. Each MoE layer has a hidden size of 64 and consists of four experts. During training, a two-layer MLP gating network produces expert selection weights via a softmax, enabling joint optimization of all experts. At inference time, we apply top-\(k\) routing based on scenario complexity and select \(k \in \{2,3,4,5,6\}\) experts. The selected expert encodings are passed through per-expert two-layer MLP heads (hidden size 64) to estimate trajectories \(\hat{y}\). Each trajectory is modeled with a Laplace distribution to capture aleatoric uncertainty in fine-grained behaviors.

\begin{table*}[htbp]
\centering
\caption{ Evaluation results on the nuScenes. \textbf{Bold} and \underline{underlined} values represent the best and second-best results.}
\setlength{\tabcolsep}{2.5mm}
\resizebox{0.85\linewidth}{!}{
\begin{tabular}{c|ccccccc}
\bottomrule
Method & Model &Publication & {minADE$_5$} $\downarrow$ & {minADE$_{10}$} $\downarrow$ & {MR$_5$} $\downarrow$ & {MR$_{10}$} $\downarrow$   \\
\hline

\multirow{9}{*}{Without LLM} 

 & AgentFormer \cite{yuan2021agentformeragentawaretransformerssociotemporal} & ICCV  & 1.86 & 1.45 & - & -\\
 
 & AFormer-FLN\cite{xu2024adaptinglengthshiftflexilength} & CVPR & 1.83 &  1.32 & - & - \\
 
 & STGM \cite{zhong2022stgm} & IEEE-TITS& - & 1.62 & -& - \\
 
 & EMSIN\cite{ren2024emsin} & IEEE-TFS & 1.77 & - & 0.54 & - \\

 & E-V$^\text{2}$-Net-SC \cite{wong2024socialcirclelearninganglebasedsocial} & CVPR & 1.44 & 1.13 & - & - \\

& GOHOME \cite{gilles2021gohomegraphorientedheatmapoutput} & ICRA& 1.42 & 1.15 & 0.57 & 0.47  \\
 & SeFlow \cite{zhang2024seflow}  & ECCV  & 1.38 & 0.98 & 0.60 & 0.38 \\
 & AutoBots \cite{girgis2022latent} & ICLR & 1.37 & 1.03 & 0.62 & 0.44  \\

\hline

 \multirow{4}{*}{LLM-based}  

    & SHIFT \cite{manas2025shift} & RSS & 2.12 & - & - & - \\
     
    & VisionTrap \cite{moon2024visiontrap} & ECCV & - & 1.48 & - & 0.56 \\

    & CoT-Drive \cite{liao2025cot} & IEEE-TAI & 1.56 & - & 0.52 & - \\

    & Traj-LLM \cite{lan2024traj} & IEEE-TIV & 1.24 & 0.99 & \textbf{0.41} & \textbf{0.23} \\

\hline
     \multirow{5}{*}{LLM + MOE} 
 
 & \textbf{Ours (2 experts)} & -  & {1.27} & {1.07} & {0.55} & 0.46 \\
 & \textbf{Ours (3 experts)} & - & \underline{1.19} & 0.97 &0.50 & 0.39 \\
 & \textbf{Ours (5 experts)}& - & 1.24 & 1.02 & 0.49 & 0.42 \\
 & \textbf{Ours (6 experts)}& - & 1.20 & \underline{0.95} & \underline{0.48} & 0.40 \\
 & \textbf{Ours (4 experts)}& - & \textbf{1.17} & \textbf{0.94} & \underline{0.48} & \underline{0.34} \\
\hline
\toprule
\end{tabular}
}
\label{nuScenes}
\end{table*}

\subsection{Experiment Results}
\subsubsection{Evaluation Results on Four Real-world Datasets}

Our model's performance is assessed using multiple real-world datasets, consistently achieving notable results across all evaluations. Specifically, on the nuScenes dataset—well-known for its complex urban scenarios—our model outperforms all other models in every metric. In particular, our model in the metrics, including \(\text{minADE$_5$}\) and \(\text{MR$_5$}\) outperform other methods by at least 13.7\% and 14.6\%, respectively, as detailed in Table \ref{nuScenes}. Remarkably, our model ranks within the top 5 on the official nuScenes leaderboard for the approach without using data augmentation via ensemble or transfer learning. Moreover, Table \ref{NGSIM/HighD/MoCAD} provides a comparison of our model's performance with others in highway, campus, and urban settings typical of right-hand driving systems. On the NGSIM dataset, for short-term predictions (1-2 seconds), our model achieves SOTA performance, comparable to the best existing methods. Its advantages are clearer in long-term predictions (3-5 seconds), where it outperforms the second-best baseline by no less than 8.7\%. This advantage is due to the effectiveness of our proposed temporal tokenizer, which improves the model's capability to capture long-term temporal dependencies in vehicle trajectories. Moreover, on the HighD dataset, our model shows an absolute advantage, outperforming other models in both short-term and long-term predictions. This success underscores the adaptability of our approach to diverse highway scenarios. 
Moreover, our model shows remarkable performance on the MoCAD dataset, further highlighting its capability to adapt to different traffic settings, ranging from unstructured roads to well-organized intersections.
In addition, our model achieves impressive results on the MoCAD dataset, further demonstrating its ability to generalize across different traffic environments, from regularly structured roads to highly irregular traffic scenes.

Collectively, these results illustrate our model's competitive performance and robustness across a wide range of traffic scenes, including urban environments, multi-agent situations, highways, and complex driving conditions. Our model demonstrates strong adaptability and scalability, making it suitable for practical deployment in AD systems.

\begin{table}[htbp]
  \centering
     \caption{{Evaluation results for our model and the other SOTA baselines in the NGSIM, HighD, and MoCAD datasets. RMSE (m) is the evaluation metric. \textbf{Bold} and \underline{underlined} values represent the best and second-best results, respectively.}}
   \setlength{\tabcolsep}{7mm}
   \resizebox{0.9\linewidth}{!}{
    \begin{tabular}{c|cccccc}
    \bottomrule
    \multicolumn{1}{c}{\multirow{2}[4]{*}{Dataset}} & \multirow{2}[4]{*}{Model} & \multicolumn{5}{c}{Prediction Horizon (s)} \\
\cmidrule{3-7}    \multicolumn{1}{c}{} &       & 1     & 2     & 3     & 4     & 5  \\
     \hline
    \multirow{10}[3]{*}{NGSIM} 
     & WSiP \cite{wang2023wsip} & 0.56  & 1.23  & 2.05  & 3.08  & 4.34  \\

    & STDAN \cite{chen2022intention} & \textbf{0.39}  & 0.96  & 1.61  & 2.56  & 3.67 \\ 
   &  DRBP\cite{gao2023dual}& 1.18  & 2.83  & 4.22  & 5.82  & - \\
   &  NLS-LSTM \cite{messaoud2019non}& 0.56  & 1.22  & 2.02  & 3.03  & 4.30  \\
   &  CF-LSTM \cite{xie2021congestion}& 0.55  & 1.10  & 1.78  & 2.73  & 3.82  \\
    &  FHIF \cite{zuo2023trajectory} &\underline{0.40}  & 0.98  & 1.66  & 2.52  & 3.63\\ 
    & HTPF \cite{hybrid} & 0.49  & 1.09  & 1.78  & 2.62  & 3.65 \\
   &  iNATran \cite{chen2022vehicle} & \textbf{0.39}  &0.96  & 1.61  & 2.42  & 3.43  \\
   &  DACR-AMTP \cite{cong2023dacr}& 0.57  & 1.07  & 1.68  & 2.53  & {3.40} \\ 
   & GaVa \cite{liao2024human} & \underline{0.40} & \underline{0.94} & \underline{1.52} & \underline{2.24} & \underline{3.13} \\
   &  \textbf{Ours} & \textbf{0.39}  & \textbf{0.90}  & \textbf{1.42}  & \textbf{2.05}  & \textbf{2.88}   \\
    \hline
    \multirow{9}[2]{*}{HighD} 
    & MHA-LSTM \cite{messaoud2021attention} & 0.19  & 0.55  & 1.10  & 1.84  & 2.78   \\
    & WSiP \cite{wang2023wsip} & 0.20  & 0.60  & 1.21  & 2.07  & 3.14  \\
    & DRBP \cite{gao2023dual} & 0.41  & 0.79  & 1.11  & 1.40  & -     \\
    & EA-Net \cite{cai2021environment} & 0.15  & 0.26  & 0.43  & 0.78  & 1.32  \\
    & DACR-AMTP \cite{cong2023dacr}& 0.10 & 0.17  & 0.31  & 0.54  & 1.01 \\ 
    & STDAN \cite{chen2022intention} & 0.19  & 0.27  & 0.48  & 0.91  & 1.66   \\
        &GaVa \cite{liao2024human}& 0.17  & 0.24  & 0.42  & 0.86  & 1.31  \\ 
    & HLTP \cite{10468619} & \underline{0.09} & \underline{0.16} & \underline{0.29} & \underline{0.38} & \underline{0.59} \\
    & \textbf{Ours} & \textbf{0.05} & \textbf{0.09} & \textbf{0.12} & \textbf{0.16} & \textbf{0.47}  \\
    \hline
    \multirow{7}[2]{*}{MoCAD} 

    & MHA-LSTM \cite{messaoud2021attention} & 1.25  & 1.48  & 2.57  & 3.22  & 4.20   \\
    & CF-LSTM \cite{xie2021congestion}& 0.72  & 0.91  & 1.73  & 2.59  & 3.44  \\
    & NLS-LSTM \cite{messaoud2019non} & 0.96  & 1.27  & 2.08  & 2.86  & 3.93 \\
    & WSiP \cite{wang2023wsip} & 0.70  & 0.87  & 1.70  & 2.56  & 3.47  \\
    & STDAN \cite{chen2022intention} & {0.62}  & {0.85}  & {1.62}  & {2.51}  & {3.32}  \\
    & HLTP \cite{10468619} & {0.55} & {0.76} & {1.44} & {2.39} & {3.21} \\
    & BAT \cite{liao2024bat}& \underline{0.35} & \underline{0.74}  & \underline{1.39}  & \underline{2.19} &\textbf{2.88}\\
    & \textbf{Ours} & \textbf{0.34} & \textbf{0.66} & \textbf{1.04} & \textbf{1.96} & \underline{2.89} \\
    \toprule
    \end{tabular}
    }
  \label{NGSIM/HighD/MoCAD}%
\end{table}%

\begin{table*}[htbp]
  \caption{{  {Comparison of our model's performance with other models on the nuScenes-corner dataset in scenarios involving turning, crowding, sudden braking, and acceleration. \textbf{Bold} and \underline{underlined} indicate the best and second-best result for each metric.}}}
  \centering
    \setlength{\tabcolsep}{2mm}
  \resizebox{0.95\linewidth}{!}
  {
  \begin{tabular}{c|ccccccccc}
    \bottomrule
    \multicolumn{1}{c}{\multirow{2}[4]{*}{Method}} & \multirow{2}[4]{*}{Model} & \multicolumn{2}{c}{Turning} & \multicolumn{2}{c}{Congested} & \multicolumn{2}{c}{Braking} & \multicolumn{2}{c}{Acceleration} \\
    
    \cmidrule(r){3-4}  \cmidrule(r){5-6}  \cmidrule(r){7-8}   \cmidrule(r){9-10}
    
    \multicolumn{1}{c}{} & & minADE$_{5}$  & minADE$_{10}$ & minADE$_{5}$  & minADE$_{10}$ & minADE$_{5}$  & minADE$_{10}$ & minADE$_{5}$  & minADE$_{10}$  \\
    \cmidrule{1-10}
    \multirow{8}{*}{Without LLM} 
    & GOHOME \cite{gilles2021gohomegraphorientedheatmapoutput} & 1.95 & 1.56 & 1.78 & 1.40 & 1.83 & 1.41 & 1.79 & 1.34 \\
    & AgentFormer \cite{yuan2021agentformeragentawaretransformerssociotemporal} & 1.89 & 1.48 & 1.73 &  1.33& 1.79 & 1.38 & 1.86 & 1.46\\
    & AutoBots \cite{girgis2022latent} & 1.80 & 1.40 & 1.28 & 1.10 & 1.26 & 1.09 & 1.31 & 1.16\\
    & NEST \cite{wang2025nest} & 1.75 & 1.49 & 1.31 & 1.05 & 1.24 & 1.02 & 1.25 & 1.10\\ 
    & PGP \cite{deo2021multimodal} & 1.73 & 1.32 & 1.19 & 0.90 & 1.19 & 0.97 & 1.23 & 0.91 \\
    & LAformer \cite{liu2024laformer} & 1.68 & 1.46 & 1.16 & 0.95 & 1.16 & 0.92 & 1.18 & 0.95 \\

    & Q-EANet \cite{chen2023q} & 1.55 & 1.43 & 1.11 & 0.98 & 1.11 & 0.99 & 1.14 & 1.04\\ 
    & DEMO \cite{wang2025dynamics} & 1.52 & 1.39 & 1.26 & 1.08 & 1.14 & 0.95 & 1.10 & \underline{0.87}\\ 
    \hline
    \multirow{5}{*}{LLM-based}
    & CoT-Drive (TinyLlama) \cite{liao2025cot} & 1.57 & 1.41 & 1.29 & 1.33 & 1.20 & 1.04 & 1.22 & 1.00 \\
    & CoT-Drive (Qwen 1.5) \cite{liao2025cot} & 1.48 & 1.33 & 1.16 & 1.24 & 1.17 & 1.00 & 1.12 & 0.94 \\
    & LMTraj-ZERO \cite{bae2024can} & 1.45 & 1.29 & 1.30 & 1.16 & 1.18 & 1.02 & 1.21 & 1.04 \\
    & LMTraj-SUP \cite{bae2024can} & 1.40 & \underline{1.19} & 1.27 & 1.10 & 1.10 & 0.98 & 1.14 & 0.93 \\ 
    & \textbf{Ours (-MoE)} & {1.86} & {1.62} & {1.52} & {1.21} & {1.64} & {1.32} & {1.55} & {1.25} \\
    \hline
    \multirow{5}{*}{LLM+MoE}

& \textbf{Ours (2 experts)} & {1.46} & {1.36} & {1.24} & {0.97} & {1.32} & {1.00} & {1.21} & {0.92} \\
  & \textbf{Ours (3 experts)} & {1.43} & {1.24} & \underline{1.06} & {0.86} & {1.16} & 0.91 & {1.14} & {0.89} \\
   & \textbf{Ours (5 experts)} & \underline{1.40} & 1.21 & {1.13} & \underline{0.84} & \underline{1.09} & {0.95} & \underline{1.08} & \underline{0.87} \\
   
  & \textbf{Ours (6 experts)} & 1.45 & 1.28 & 1.08 & 0.89 & 1.13 & \underline{0.90} & 1.13 & 0.92 \\

    & \textbf{Ours (4 experts)} & \textbf{1.38} & \textbf{1.18} & \textbf{1.03} & \textbf{0.83} & \textbf{1.05} & \textbf{0.86} & \textbf{1.03} & \textbf{0.84} \\
\toprule
  \end{tabular}
  }
   \label{longtailed-results}
\end{table*}

{ 
\subsubsection{Evaluation Results for Computational Performance}
To assess the efficiency of our model, we benchmark inference latency on the nuScenes dataset under the same setting as Traj-LLM (RTX~4090, batch size $=12$), reporting average per-scene time for 12 agents. Table~\ref{computational performance} summarizes the results and includes baseline figures from Traj-LLM~\cite{lan2024traj}. For a fair comparison, WM-MoE is evaluated by computing its mean latency over all scenes with 12 agents, yielding the values reported.
The results clearly demonstrate that, compared to TrajLLM, which also utilizes GPT 2 as its LLM, our model achieves significantly lower inference times. Relative to PGP (215\,ms, 1.30 ADE), WM-MoE achieves a \(3.36\times\) speedup (\(\downarrow70.2\%\)) with \(10.0\%\) lower ADE; versus LLM-Traj, which also utilizes a GPT-2 backbone, WM-MoE is \(1.53\times\) faster (\(\downarrow34.7\%\)) while concurrently achieving a  \(5.6\%\) lower ADE. The efficiency gains are even more pronounced when compared with other baselines. Compared with DEMO, our model achieves a \(2.64\times\) speedup (62.1\% faster) with 2.5\% lower ADE. It also runs 1.80 times faster than LAformer with 1.7\% lower ADE. Notably, WM-MoE uses only 0.12M trainable parameters, 71\%–98\% fewer than the baselines, yielding a favourable accuracy–efficiency trade-off.
Consequently, WM-MoE's performance meets the real-time processing requirements of typical Level-3 systems and is well-suited for deployment on contemporary automotive SoCs like Tesla FSD (254 TOPS), NVIDIA Orin (254 TOPS), and NVIDIA DRIVE series (500-700 TOPS), with ample headroom for concurrent perception and planning tasks.}

\begin{table}[!t]
\caption{Computational performance of WM-MoE and SOTA baseline models on the nuScenes dataset. \textbf{Bold} values represent the best results.}
  \centering
  \setlength{\tabcolsep}{2mm}
  \resizebox{0.7\linewidth}{!}{
    \begin{tabular}{ccccc}
    \bottomrule
      \multirow{1}{*}{Model}  & \multicolumn{1}{c}{minADE\(_{5}\)} & \multicolumn{1}{c}{Trainable Params} & \multicolumn{1}{c}{Batch Size} &  \multicolumn{1}{c}{Inference Speed} \\
      \hline
     PGP \cite{deo2021multimodal} & 	1.30	& 0.42M	& 12& 	215ms\\
     Traj-LLM \cite{lan2024traj} & 	1.24	& 7.72M & 	12	& 98ms\\
     DEMO \cite{wang2025dynamics} & 	1.20	& 0.87M & 12	& 169ms\\
     LAformer \cite{liu2024laformer} & 1.19 & 2.59M & 12 & 115ms\\
     \hline
     \textbf{Ours} & \textbf{1.17}	&\textbf{0.12M}& 	12& 	\textbf{64ms}\\
      \toprule
    \end{tabular}
    }
    \label{computational performance}
\end{table}

{ 
\subsubsection{Evaluation Results on the nuScenes-corner Dataset}
We evaluate the performance of our model and several top baselines on the nuScenes-corner dataset, specifically focusing on corner-case scenarios including turning maneuvers, traffic congestion, sudden braking, and sudden acceleration. As shown in Table \ref{longtailed-results}, our model demonstrates a significant advantage over the baseline models in addressing these challenging scenarios. In turning, congested, braking, and acceleration cases, our model shows substantial improvements in \(\text{minADE}_5\) and \(\text{minADE}_{10}\), outperforming baselines by at least 5.7\% and 15.1\%, respectively. Moreover, a direct comparison among LLM-based methods reveals the architectural superiority of our WM-MoE framework. While methods like LMTraj-SUP show competitive performance, our optimal 4-expert model (Ours (4 experts)) consistently achieves lower error across the board. The advantage is particularly pronounced in Congested scenarios, where our model achieves a \(\text{minADE}_5\) of 1.03, a significant 18.9\% improvement over LMTraj-SUP's 1.27. This suggests that while LLMs provide valuable high-level context, our model's ability to specialize with its MoE module is critical for disentangling the complex, low-speed interactions inherent in dense traffic. The most crucial insight, however, comes from the ablation study Ours (-MoE). This model, which uses our Language-Enhanced Encoder but removes the MoE decoder, serves as a direct point of comparison. Its performance is dramatically worse than our full model and is even inferior to many non-LLM baselines. For example, in the Turning scenario, the  \(\text{minADE}_5\) degrades from 1.38 to 1.86 (a 34.8\% increase in error) without the MoE network. This result unequivocally demonstrates that simply incorporating an LLM is insufficient; the performance of WM-MoE is driven by the synergistic combination of the LLM's contextual priors and the MoE's capacity to apply specialized, fine-grained reasoning to specific scenario dynamics. In addition, the results within the LLM and MoE category validate our choice of a 4-expert configuration, as it consistently delivers the best or second-best performance across all metrics. Using fewer experts (2 or 3) or more (5 or 6) leads to a marginal decline in accuracy, suggesting that four experts provide an optimal balance between specialized knowledge and model complexity for this task. These results indicate that our MoE architecture is particularly effective in addressing the limitations observed in previous models when handling these difficult scenarios. The specialized experts in the MoE are capable of adapting to the unique demands of each corner case, significantly enhancing the model's robustness and accuracy in safety-critical conditions. Furthermore, WM-MoE, utilizing MoE networks, significantly outperforms leading models in \(\text{minADE}_5\) and \(\text{minADE}_{10}\), demonstrating the potential of the MoE-based model in handling these high-risk situations.}

Overall, these findings sparked our interest in exploring the impact of varying the number of expert models on overall performance. In our ablation study, we analyze how the number of experts influences model adaptability and efficiency in handling diverse, challenging traffic scenarios.

}

\begin{table}[htbp]
   \caption{Ablation results of different LLM models on the nuScenes testset. \textbf{Bold} values represent the best results in each category.}
  \centering
  \setlength{\tabcolsep}{2mm}
  \resizebox{0.6\linewidth}{!}{
    \begin{tabular}{ccccc}
    \bottomrule
      \multirow{2}{*}{Datasets}  & \multicolumn{4}{c}{Metrics} \\ \cmidrule{2-5}
      & minADE\(_5\) & minADE\(_{10}\) & MR\(_5\) & MR\(_{10}\) \\ \midrule
     nuScenes (drop 1 frame)& 	1.25	& 1.06	& 0.51& 	0.48\\
nuScenes (drop 2 frames)& 	1.34	& 1.14& 	0.56	& 0.52\\
nuScenes (drop 3 frames)& 	1.39	& 1.17	& 0.59	& 0.54\\
nuScenes & \textbf{1.17}	&\textbf{0.94}& 	\textbf{0.48}& 	\textbf{0.34}\\
      \toprule
    \end{tabular}
    }
    \label{missing-data}
\end{table}

{ 
\subsubsection{Evaluation Results on the Data-Missing Scenes} 
A notable gap in current motion forecasting methodologies is their dependence on the entire historical trajectory to forecast future trajectories. However, these approaches often overlook critical factors inherent in real-world driving conditions. The main challenges arise from observational limitations, including sensor constraints and environmental factors such as obstacles, adverse weather, or traffic congestion. Most existing models \cite{liao2024bat,lan2024traj,chen2023q} are trained and evaluated on datasets with complete or perfect observations, which leads to a marked decline in performance when faced with missing data. Furthermore, such data deficiency scenarios are rarely addressed in datasets but have substantial implications for driving safety. To our knowledge, missing data in motion forecasting has been extensively studied for pedestrians but remains largely unaddressed for AVs.

To address this limitation, we explicitly assess the model’s robustness under varying degrees of input sparsity, simulating realistic scenarios where perception data is partially missing or degraded. As shown in Table \ref{missing-data}, our model consistently demonstrates strong predictive performance despite incomplete observations. Specifically, on the nuScenes (drop 1 frame) dataset, our model outperforms nearly all SOTA baselines that were trained on complete data, demonstrating significantly lower values in the \(\text{minADE}\), \(\text{minFDE}\), and \(\text{MR}\) metrics—highlighting its superior predictive capabilities despite missing data. When evaluated on the nuScenes (drop 2 frames) dataset, the model's accuracy declines due to the increased data loss, as expected. However, it still surpasses competitive baselines such as GOHOME, E-V$^\text{2}$-Net-SC, and SeFlow, underscoring its robustness in handling partially missing inputs. For the most challenging scenario, with 60\% of data missing, the prediction performance naturally deteriorates as more input frames are absent. Remarkably, even under these severe conditions, our model maintains a substantial advantage, with evaluation metrics remaining slightly better than the majority of the top 30 ranked baselines. These results emphasize the model’s resilience and adaptability, underscoring its broad practical applicability in real-world autonomous driving systems where sensor data can be unreliable or incomplete.}

{ 
\subsubsection{Evaluation Results of Data Imbalance}

A fundamental challenge for motion forecasting models is the inherent long-tail distribution of real-world driving data. To rigorously evaluate the robustness of WM-MoE to this data imbalance, we design a controlled experiment using the nuScenes and nuScenes-corner datasets. We first curate a balanced test set of 250 samples, evenly distributed across five categories: Turning, Congested, Braking, Acceleration, and Common driving scenes. Subsequently, we construct five training sets with progressively skewed distributions, ranging from a ``Common-Only'' set with zero corner-case examples to increasingly balanced sets by reducing the number of common scenarios from ~46k down to 1k. This setup allows us to systematically quantify how performance on rare events is affected by their prevalence during training. The following is the distribution of five specific situations:
\begin{figure}
\centering
\includegraphics[width=0.9\textwidth]{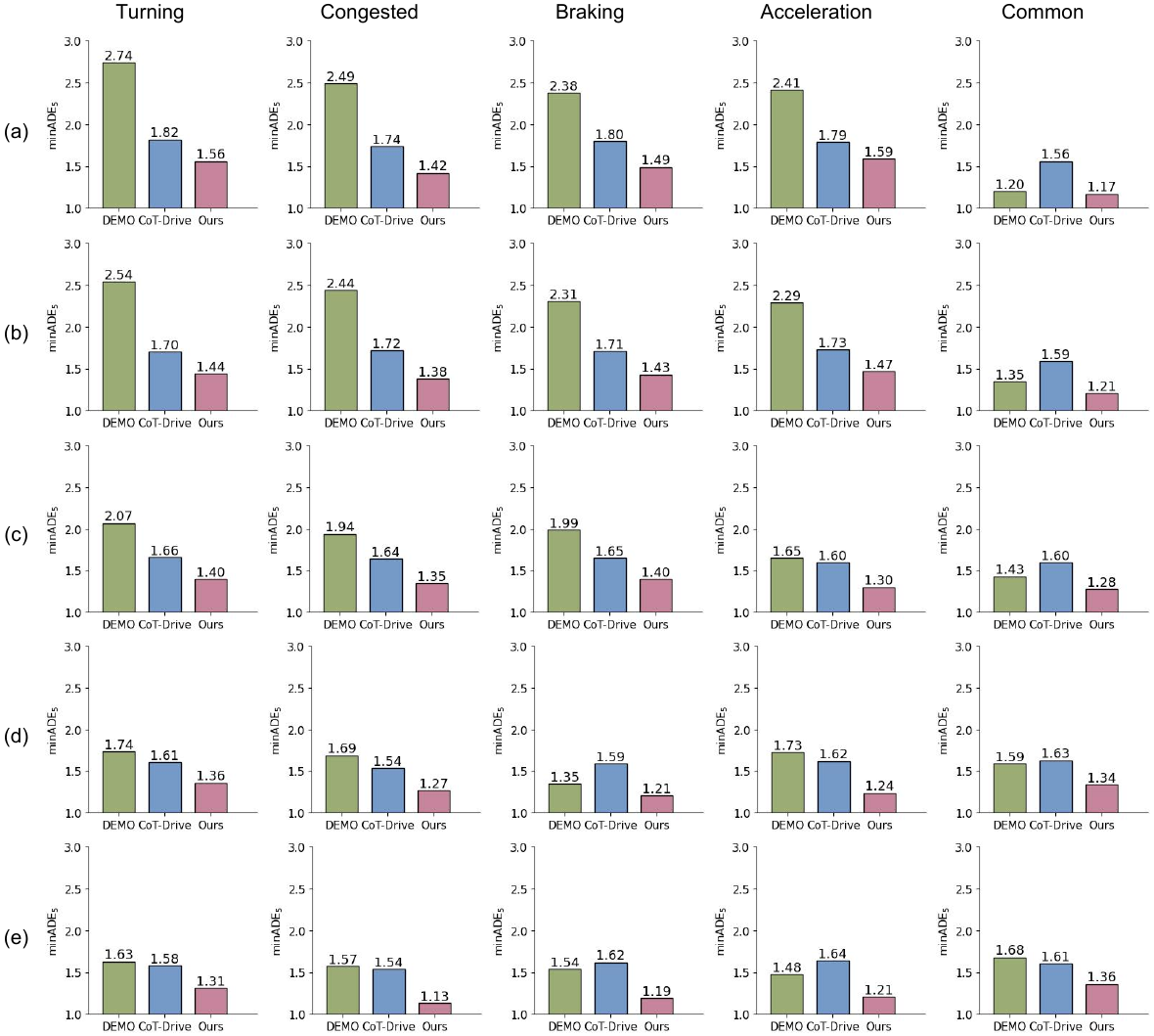}
\caption{Comparative performance of our proposed WM-MoE against baselines (CoT-Drive and DeMo) on a balanced 250-scene test set (50 per class), evaluated using minADE$_5$. Columns report performance on Turning, Congested, Braking, Acceleration, and Common scenarios. Rows (a–e) correspond to the five training splits with different head–tail imbalance: (a) Common-only, (b) Full dataset, (c) Reduced-Common 20k, (d) Reduced-Common 5k, and (e) Reduced-Common 1k.}
\label{fig:imbalance}
\end{figure}
\begin{itemize}
    \item (a) Common-Only: Comprises 46,345 Common scenes, excluding corner-case scenarios.
    \item (b) Complete Dataset: Includes 1,070 Turning, 934 Congested, 782 Braking, 406 Accelerating, and 46,345 Common scenes.
    \item (c) Reduced Common (20,000): Includes 1,070 Turning, 934 Congested, 782 Braking, 406 Accelerating, and 20,000 Common scenes.
    \item (d) Reduced Common (5,000): Includes 1,070 Turning, 934 Congested, 782 Braking, 406 Accelerating, and 5,000 Common scenes.
    \item (e) Reduced Common (1,000): Includes 1,070 Turning, 934 Congested, 782 Braking, 406 Accelerating, and 1,000 Common scenes.
\end{itemize}

We benchmark WM-MoE against a strong non-LLM baseline (DeMo) and an alternative LLM-based method (CoT-Drive) in these conditions, with results summarized in Figure \ref{fig:imbalance}. Our empirical results indicate that our model consistently attains SOTA performance across all tested datasets. Specifically, in the most extreme case (row a), where models are trained exclusively on common scenarios, WM-MoE demonstrates a remarkable ability to generalize to all four unseen corner-case categories. It significantly outperforms both the non-LLM baseline and CoT-Drive. This suggests that the pre-trained priors, accessed via our language-enhanced world model, provide a foundational understanding of dynamic interactions that transcends the immediate training data. Moreover, as the training data becomes more balanced (from row b to row e), all models improve their performance on corner cases, as expected. However, WM-MoE consistently maintains the lowest error across all five scenario types and all five data distributions. The performance gap is most pronounced in the highly imbalanced regimes, highlighting our model's superior data efficiency. Notably, the consistent outperformance of WM-MoE over CoT-Drive, another LLM-enhanced model, underscores the efficacy of our specific architectural design. While both models benefit from LLM priors, the superior performance of WM-MoE is attributable to its MoE network, which provides a principled mechanism to ``divide and conquer'' the problem space. By routing scenarios to specialized experts, our model can learn the nuanced dynamics of rare events without compromising its accuracy on common scenarios. 

To sum up, these evaluation results furnish robust evidence advocating for the incorporation of LLMs into trajectory prediction models as a potent approach to addressing the inherent challenges of data imbalance.}

\subsection{Ablation Studies}
\subsubsection{Importance of Each Component}
 Understanding the contributions of different components within a model and their interactions is crucial for refining and enhancing motion forecasting performance. Therefore, as shown in Table \ref{ablation-component} , we conduct extensive ablation studies on the nuScenes and nuScenes-corner datasets to determine the unique impact of each component in our proposed model. These studies allow us to understand better how each part contributes to overall performance and to identify which elements are most critical for robustness and accuracy in complex driving scenarios. Specifically, the ablation models are defined below, each compared against Model E, which incorporates all components:
\begin{itemize}
    \item Model A: Omits the MoE model in the Decision Module.
    \item Model B: Removes the Language-Enhanced Encoder in the Memory Module.
    \item Model C: Excludes the BEV input in this model.
    \item Model D: Removes the temporal tokenizer in the Language-Enhanced Encoder.
    \item Model E: Full model proposed in this study.
\end{itemize}
\begin{table}[htbp]
  \centering
    \caption{Ablation study for different components of our proposed WM-MoE model.}
  \label{ablation-component}
  \setlength{\tabcolsep}{4mm}
  \resizebox{0.55\linewidth}{!}{
    \begin{tabular}{cccccccc}
      \bottomrule
      \multirow{2}[3]{*}{Component} & \multicolumn{4}{c}{Ablation Model} \\ 
      \cmidrule(r){2-6} 
      & A & B & C & D & E \\ 
      \midrule
      MoE Model & \ding{55} & \ding{52} & \ding{52} & \ding{52} &\ding{52} \\ 
      Language-Enhanced Encoder & \ding{52}& \ding{55} & \ding{52}  & \ding{52} & \ding{52}\\ 
      BEV Input & \ding{52} & \ding{52}& \ding{55} & \ding{52} & \ding{52}\\ 
      Temporal Tokenizer & \ding{52} &\ding{52} & \ding{52} &\ding{55} &\ding{52} \\
      \toprule
    \end{tabular}
  }
\end{table}

\begin{table}[htbp]
  \centering
   \caption{Ablation results on the nuScenes dataset. \textbf{Bold} values represent the best results.}
  \setlength{\tabcolsep}{4mm}
  \resizebox{0.8\linewidth}{!}{
    \begin{tabular}{ccccccc}
    \bottomrule
      \multirow{2}{*}{Dataset} & \multirow{2}{*}{Metrics} & \multicolumn{4}{c}{Model} \\ \cmidrule{3-7}
      &  & A & B & C & D & E\\ 
      \hline
      \multirow{4}{*}{nuScenes} 
      & minADE$_{\text{5}}$ & 1.41$_{\downarrow 20.5\%}$ & 1.38$_{\downarrow 18.0\%}$ & 1.32$_{\downarrow 12.8\%}$ & 1.31$_{\downarrow 12.0\%}$ & \textbf{1.17}  \\
      & minADE$_{\text{10}}$ & 1.32$_{\downarrow 40.4\%}$ & 1.25$_{\downarrow 33.0\%}$ & 1.17$_{\downarrow 24.5\%}$ & 1.18$_{\downarrow 25.5\%}$ & \textbf{0.94}  \\
      & MR$_{\text{5}}$ & 0.71$_{\downarrow 47.9\%}$ & 0.62$_{\downarrow 29.2\%}$ & 0.60$_{\downarrow 25.0\%}$ & 0.57$_{\downarrow 18.8\%}$ & \textbf{0.48}   \\
      & MR$_{\text{10}}$ & 0.62$_{\downarrow 82.4\%}$ & 0.58$_{\downarrow 70.6\%}$ & 0.54$_{\downarrow 58.8\%}$  & 0.55$_{\downarrow 61.8\%}$ & \textbf{0.34}  \\
      \toprule
    \end{tabular}
    }

    \label{ablation-result}
\end{table}

\begin{table*}[htbp]
  \centering
      \caption{{ Ablation results on the nuScenes-corner dataset. \textbf{Bold} values represent the best results.}}
    \label{ablation-corner}
  \setlength{\tabcolsep}{4mm}
  \resizebox{\linewidth}{!}{
    \begin{tabular}{ccccccccc}
    \bottomrule
      \multirow{2}{*}{Model} & \multicolumn{2}{c}{Turning} & \multicolumn{2}{c}{Congested} & \multicolumn{2}{c}{Braking} & \multicolumn{2}{c}{Acceleration} \\  
      
      \cmidrule(r){2-3} \cmidrule(r){4-5}  \cmidrule(r){6-7} \cmidrule(r){8-9}
       &  minADE$_{5}$  & minADE$_{10}$ & minADE$_{5}$  & minADE$_{10}$ & minADE$_{5}$  & minADE$_{10}$ & minADE$_{5}$  & minADE$_{10}$  \\
    \midrule
        A & 1.73 $_{\downarrow 25.4\%}$ & 1.45 $_{\downarrow 22.9\%}$ & 1.51 $_{\downarrow 31.1\%}$ & 1.25 $_{\downarrow 33.6\%}$ & 1.33 $_{\downarrow 26.7\%}$ & 1.21 $_{\downarrow 28.9\%}$ & 1.36 $_{\downarrow 32.0\%}$ & 1.24 $_{\downarrow 32.3\%}$ \\
        B & 1.68 $_{\downarrow 21.7\%}$ & 1.39 $_{\downarrow 17.8\%}$ & 1.34 $_{\downarrow 23.3\%}$ & 1.12 $_{\downarrow 25.9\%}$ & 1.27 $_{\downarrow 20.9\%}$ & 1.12 $_{\downarrow 23.2\%}$ & 1.28 $_{\downarrow 24.3\%}$ & 1.17 $_{\downarrow 28.0\%}$ \\
        C & 1.63 $_{\downarrow 18.1\%}$ & 1.32 $_{\downarrow 11.9\%}$ & 1.27 $_{\downarrow 18.9\%}$ & 1.08 $_{\downarrow 22.2\%}$ & 1.19 $_{\downarrow 13.3\%}$ & 1.07 $_{\downarrow 19.5\%}$ & 1.24 $_{\downarrow 20.4\%}$ & 1.06 $_{\downarrow 20.2\%}$ \\
        D & 1.59 $_{\downarrow 15.2\%}$ & 1.27 $_{\downarrow 7.6\%}$ & 1.15 $_{\downarrow 11.7\%}$ & 0.97 $_{\downarrow 14.4\%}$ & 1.14 $_{\downarrow 8.3\%}$ & 0.99 $_{\downarrow 13.1\%}$ & 1.17 $_{\downarrow 13.6\%}$ & 0.98 $_{\downarrow 14.3\%}$ \\   
        E & \textbf{1.38} & \textbf{1.18} & \textbf{1.03} & \textbf{0.83} & \textbf{1.05} & \textbf{0.86} & \textbf{1.03} & \textbf{0.84} \\
    \toprule
    \end{tabular}
    }
\end{table*}

{As shown in Table \ref{ablation-result} and Table \ref{ablation-corner}, the comparative analysis on the nuScenes and nuScenes-corner datasets demonstrates the substantial contributions of each key module in our model. Notably, Model E, which incorporates all components, consistently outperforms the other variants across all evaluation metrics, highlighting the collective importance of each component in achieving optimal performance.

Model A, which excludes the MoE module, shows significant performance drops on the nuScenes dataset, with reductions of \(\text{minADE}_{5}\), \(\text{minADE}_{10}\), and \(\text{MR}_{5}\) and \(\text{MR}_{10}\) reducing by 20.5\%, 40.4\%, 47.9\% and 82.4\%, respectively.
Furthermore, on the nuScenes-corner dataset, \(\text{minADE}_{5}\) achieves reductions of 25.4\%, 31.1\%, 26.7\%, and 32.0\% for the Turning, Congested, Braking, and Acceleration scenarios, respectively. This clearly illustrates the critical role of the MoE module in improving motion forecasting accuracy. The MoE module allows different expert networks to share features in a common space and specialize in handling specific scenarios, effectively breaking down a complex task into a set of simpler sub-tasks corresponding to sub-domains in the input space. Through our proposed novel routing mechanism, each expert can contribute precisely where needed, collaboratively aggregating critical features to tackle the challenges of complex traffic scenarios.

Moreover, Model B, which omits the Language-Enhanced Encoder, demonstrates a substantial decrease in predictive performance. This decline suggests the significant contribution of LLMs, especially in enhancing model accuracy for complex urban scenarios where a rich understanding of contextual and spatial relationships is crucial. Similarly, Model C shows a performance drop when BEV input is removed from WM-MoE, highlighting the importance of BEV in enriching data modalities and enhancing semantic understanding. Finally, Model D, which removes the temporal tokenizer within the Language-Enhanced Encoder, also shows a significant performance degradation for all the nuScenes and nuScenes-corner datasets, even when LLMs are applied. This result highlights the importance of the tokenizer in aligning time-series data with pre-trained LLM patterns learned from natural language corpora. More importantly, it helps map the fine-grained features of trajectory data to LLM-compatible modalities, enabling LLMs to effectively exploit their scene understanding capabilities and capture temporal dependencies in long-term predictions.} 

To further validate the significance of each component, we visualize the results of the ablation study in Section \ref{Ablation Models}, providing an intuitive understanding of how these components interact to optimize model performance.

\begin{table*}[htbp]
  \centering
      \caption{{  Ablation of different LLM backbones in the Language-Enhanced Encoder on the nuScenes dataset. \textbf{Bold} and \underline{underlined} indicate the best and second-best result for each metric. Values highlighted in \textcolor{blue}{blue} denote the LLM selected in WM-MoE.}}
  \resizebox{\linewidth}{!}{
    \begin{tabular}{cccccccc}
    \bottomrule

       Model& \#Param. (B)\hspace{3mm} &Inference Time (ms) & Tokenizer\hspace{3mm} & minADE\(_5\) & minADE\(_{10}\) & MR\(_5\) & MR\(_{10}\) \\ 
      \hline
         Vicuna (Lora fine-tuning) \cite{zheng2023judging}& 13.00 &  158.84   & \ding{55}& 1.47 & 1.26 & 0.60 & 0.53  \\ 
         Llama2 (Lora fine-tuning) \cite{touvron2023llama}& 7.00 &  86.92   & \ding{55}& 1.38 & 1.21 & 0.58 & 0.49  \\ 
         TinyLlama (Lora fine-tuning) \cite{zhang2024tinyllama}& 1.10 &   16.67  & \ding{55}& 1.31 & 1.16 & 0.54 & 0.55  \\ 
        GPT Neo (Lora fine-tuning) \cite{gao2020pile}&  0.34& 8.33  & \ding{55}& 1.29 & 1.17 & 0.57 & 0.54\\
        GPT 2 (Lora fine-tuning) \cite{radford2019language}& 0.11& 4.83 &  \ding{55}& 1.27 & 1.04 & 0.50 & 0.43  \\
         Vicuna (frozen) \cite{zheng2023judging}& 13.00 &  163.67   & \ding{52}& \textbf{1.14} & \textbf{0.90} & 0.50 & 0.42  \\ 
         Llama2 (frozen) \cite{touvron2023llama}& 7.00 &  92.43  & \ding{52}& \underline{1.16} & \underline{0.92} & 0.48 & \underline{0.36}  \\ 
        TinyLlama (frozen) \cite{zhang2024tinyllama}& 1.10 &  17.91 & \ding{52}& \underline{1.16} & 0.95 & \textbf{0.47} & \underline{0.36} \\

     GPT Neo (frozen) \cite{gao2020pile}& 0.34 & 9.58 & \ding{52}& 1.18 & \underline{0.92} & 0.49 & 0.39 \\
           \rowcolor{blue!8} GPT 2 (frozen) \cite{radford2019language}& 0.11	& 5.67  & \ding{52} & 1.17 & 0.94 & \underline{0.48} & \textbf{0.34} \\
    \toprule
    \end{tabular}
    }
    \label{LLM_test}
\end{table*}

\begin{table*}[htbp]
  \centering
       \caption{{  Ablation of different LLM backbones in the Language-Enhanced Encoder on the nuScenes-corner. \textbf{Bold} and \underline{underlined} indicate the best and second-best result for each metric. Values highlighted in \textcolor{blue}{blue} denote the LLM selected in WM-MoE.}}
    \label{LLM_test 1}
  \setlength{\tabcolsep}{1mm}
  \resizebox{\linewidth}{!}{
    \begin{tabular}{ccccccccccc}
    \bottomrule
      \multirow{2}{*}{Model} & \multirow{2}{*}{\#Param. (B)\hspace{3mm}} & \multirow{2}{*}{Tokenizer \hspace{3mm}} & \multicolumn{2}{c}{Turning} & \multicolumn{2}{c}{Congested} & \multicolumn{2}{c}{Braking} & \multicolumn{2}{c}{Acceleration} \\  
      
      \cmidrule(r){4-5} \cmidrule(r){6-7}  \cmidrule(r){8-9} \cmidrule(r){10-11}
       & & & minADE$_{5}$  & minADE$_{10}$ & minADE$_{5}$  & minADE$_{10}$ & minADE$_{5}$  & minADE$_{10}$ & minADE$_{5}$  & minADE$_{10}$  \\
    \hline
        Vicuna (Lora fine-tuning) \cite{zheng2023judging} & 13.00  & \ding{55}& 1.58 & 1.43 & 1.12 & 0.89 & 1.30 & 0.97 & 1.30 & 1.00 \\
        Llama2 (Lora fine-tuning) \cite{touvron2023llama}& 7.00 & \ding{55} & 1.55 & 1.42 & 1.27 & 1.05 & 1.33 & 0.99 & 1.32 & 1.05 \\    
        TinyLlama (Lora fine-tuning) \cite{zhang2024tinyllama} & 1.10  & \ding{55}& 1.55 & 1.41 & 1.22 & 0.98 & 1.38 & 1.10 & 1.26 & 0.99 \\
        GPT Neo (Lora fine-tuning) \cite{gao2020pile}& 0.34 & \ding{55} & 1.57 & 1.40 & 1.23 & 1.00 & 1.36 & 1.05 & 1.25 & 1.02 \\     
        GPT 2 (Lora fine-tuning) \cite{radford2019language}& 0.11 & \ding{55}	& 1.52 & 1.34 & 1.14 & 0.91 & 1.29 & 0.98 & 1.24 & 0.93 \\

        Vicuna (frozen) \cite{zheng2023judging} & 13.00  & \ding{52}& \textbf{1.32} & \textbf{1.14} & \textbf{1.00} & \textbf{0.83} & 1.08 & 0.92 & \underline{1.00} & \textbf{0.80} \\
        Llama2 (frozen) \cite{touvron2023llama}& 7.00 & \ding{52} & 1.37 & \underline{1.16} & 1.05 & \underline{0.84} & 1.06 & \underline{0.88} & \textbf{0.99} & \textbf{0.80} \\          
        TinyLlama (frozen) \cite{zhang2024tinyllama} & 1.10  & \ding{52}& \underline{1.35} & 1.17 & \underline{1.03} & \textbf{0.83} & \textbf{1.03} & \textbf{0.86} & 1.04 & \underline{0.82} \\
        GPT Neo (frozen) \cite{gao2020pile} & 0.34 & \ding{52}& 1.36 & 1.17 & 1.04 & 0.85 & \underline{1.05} & 0.90 & 1.07 & {0.84} \\
     \rowcolor{blue!8}    GPT 2 (frozen) \cite{radford2019language}& 0.11 & \ding{52}& 1.38 & 1.18 & \underline{1.03} & \textbf{0.83} & \underline{1.05} & \textbf{0.86} & 1.03 & {0.84}\\
        
    \toprule
    \end{tabular}
    }
\end{table*}

\begin{figure*}
  \centering
  \includegraphics[width=\linewidth]{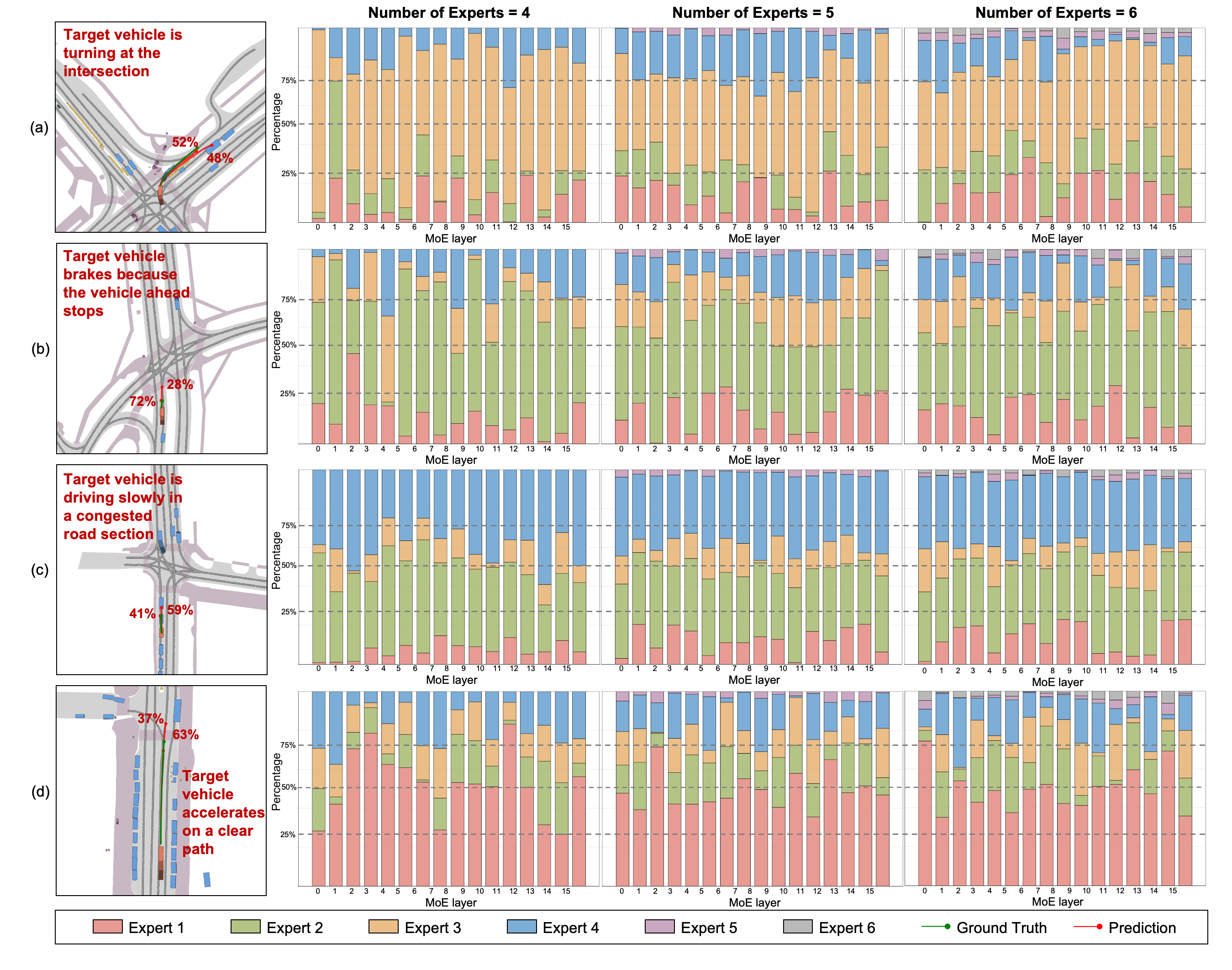}
 \caption{{Visualization of expert load distribution on the nuScenes-corner dataset. The discontinuous lines illustrate how the contributions of different experts are distributed across various MoE layers. Panels (a), (b), (c), and (d) represent the turning, braking, congested, and acceleration scenarios, respectively. Each expert specializes in different MoE layers, with new experts progressively added in higher layers. As the number of experts increases, the additional experts contribute minimally, with their impact becoming almost negligible in the higher MoE layers.}}
  \label{visualization-moe}
\end{figure*}
 {
\subsubsection{Ablation Study for LLM Backbones}

We further evaluate multiple LLM backbones within the Language-Enhanced Encoder of WM-MoE under two regimes: frozen backbones with our temporal tokenizer and LoRA-finetuned backbones without the tokenizer. Table~\ref{LLM_test} and Table~\ref{LLM_test 1} present the experimental results on the nuScenes test set and the nuScenes-corner benchmark, respectively. The results consistently demonstrate that frozen LLMs with the proposed tokenizer outperform LoRA-finetuned counterparts without the tokenizer across both common-case and corner-case scenarios. Notably, while increasing the parameter scale of LLMs can yield marginal improvements in prediction accuracy, the gains are often overshadowed by the associated computational overhead. Specifically, Vicuna achieves SOTA performance relative to other models, with a 2.56\% improvement in the \(\text{minADE}_{5}\) metric on the nuScenes dataset compared to GPT-2, and average performance gains of 0.94\% and 2.47\% for the \(\text{minADE}_{5}\) and \(\text{minADE}_{10}\) metrics, respectively, across the four scenarios of the nuScenes-corner dataset, these incremental improvements are substantially outweighed by its elevated computational demands. Specifically, Vicuna's inference latency is around 29 times higher, and its model size is approximately 118 times larger than that of GPT-2, rendering the resource requirements disproportionate to the performance benefits. Consequently, our adoption of GPT-2 underscores the efficacy of the proposed framework in achieving real-time, lightweight processing, thereby significantly enhancing its practicality for deployment in safety-critical trajectory prediction tasks. Moreover, models incorporating the temporal tokenizer without LoRA fine-tuning consistently outperform those relying solely on LoRA fine-tuning, achieving at least a 7.87\% improvement in the \(\text{minADE}_{5}\).  We speculate that LoRA may inadvertently distort internal representations of the pretrained LLM, reducing its alignment with scene dynamics and increasing the risk of overfitting. In contrast, our tokenizer enhances temporal reasoning while preserving the backbone’s structure, allowing better abstraction from trajectory inputs and mitigating the effects of catastrophic forgetting. This design not only improves generalization in both balanced and imbalanced settings but also strengthens adaptability in complex interactive scenes. These observations validate our decision to adopt frozen GPT-2 with the temporal tokenizer in WM-MoE as a robust and efficient encoder for motion forecasting.
}

{ 
\subsubsection{Ablation Study for MoE Model}

We conduct comprehensive experiments to analyze the impact of varying the number of experts in the MoE model on performance. Results are summarized in Table \ref{nuScenes} and Table \ref{longtailed-results}, which present findings on the nuScenes online test dataset and the nuScenes-corner dataset, respectively. The experiments reveal that both insufficient and excessive numbers of experts can adversely affect prediction accuracy, underscoring the importance of selecting an optimal configuration. Table \ref{nuScenes} highlights that the optimal model, consisting of four experts, shows enhancements of 1.7\% and 4.2\% in minADE$_5$ and MR$_5$, respectively, compared to the model with three experts.  Furthermore, our proposed model experiences a significant performance drop when the MoE component is removed. Interestingly, relative to the model with five experts, the optimal model with four experts shows improvements of 6.0\% and 2.1\% in minADE$_5$ and MR$_5$, respectively. These findings indicate that the number of experts directly correlates with the complexity of motion forecasting tasks. A small number of experts may miss the diversity of behaviors and interactions in traffic scenes, whereas too many can introduce redundancy, increase complexity, and promote overfitting. Choosing an appropriate expert count is therefore critical to optimizing performance and maintaining generalization to unseen scenarios.
}

{ 
\subsection{Case Study for MoE Model}
Figure \ref{visualization-moe} illustrates the distribution of contributions and expert preferences in our MoE model across different scenarios in the nuScenes-corner dataset. As we vary the number of experts, distinct patterns of task allocation emerge. When the number of experts is set to 4, as shown in panel (a) for a turning scenario, Expert 3 dominates the lower MoE blocks (0-5), handling the majority of the tokens. In the higher MoE blocks (6-15), Experts 1 and 4 begin to contribute more, while Expert 2 carries minimal to no load.

In panel (b), which depicts a sudden braking scenario, Expert 2 takes the lead. As the model deepens, the load gradually shifts to include Experts 4 and 1. Similarly, in the crowded scenario shown in panel (c), Experts 4 and 2 shoulder most of the responsibility, while the other experts contribute less. Panel (d) demonstrates a multi-agent acceleration scenario, where Experts 1 and 4 are crucial in processing the scene. These visualizations reveal that the distribution of tasks is not rigidly tied to specific experts for distinct scene categories. Instead, multiple experts collaboratively analyze the motion forecasting for different scenes. In contrast to traditional MoE models, which select a subset of top-performing experts for each input, our approach aggregates the outputs of all experts. The routing mechanism assigns weighted distributions to each expert, ensuring that the decoder benefits from the full spectrum of knowledge embedded in the experts. This design facilitates effective handling of corner cases by enabling progressive, layered collaboration and task division.

Moreover, as the number of experts increases to 5 and 6, as shown in Figure \ref{visualization-moe}, their contributions become marginal, suggesting that adding more experts does not significantly alter the task distribution. This indicates that our model achieves an optimal balance in task allocation, where adding more experts beyond a certain point does not improve performance. In fact, additional experts only introduce redundancy and computational overhead.}

\begin{figure*}
  \centering
  \includegraphics[width=0.9\linewidth]{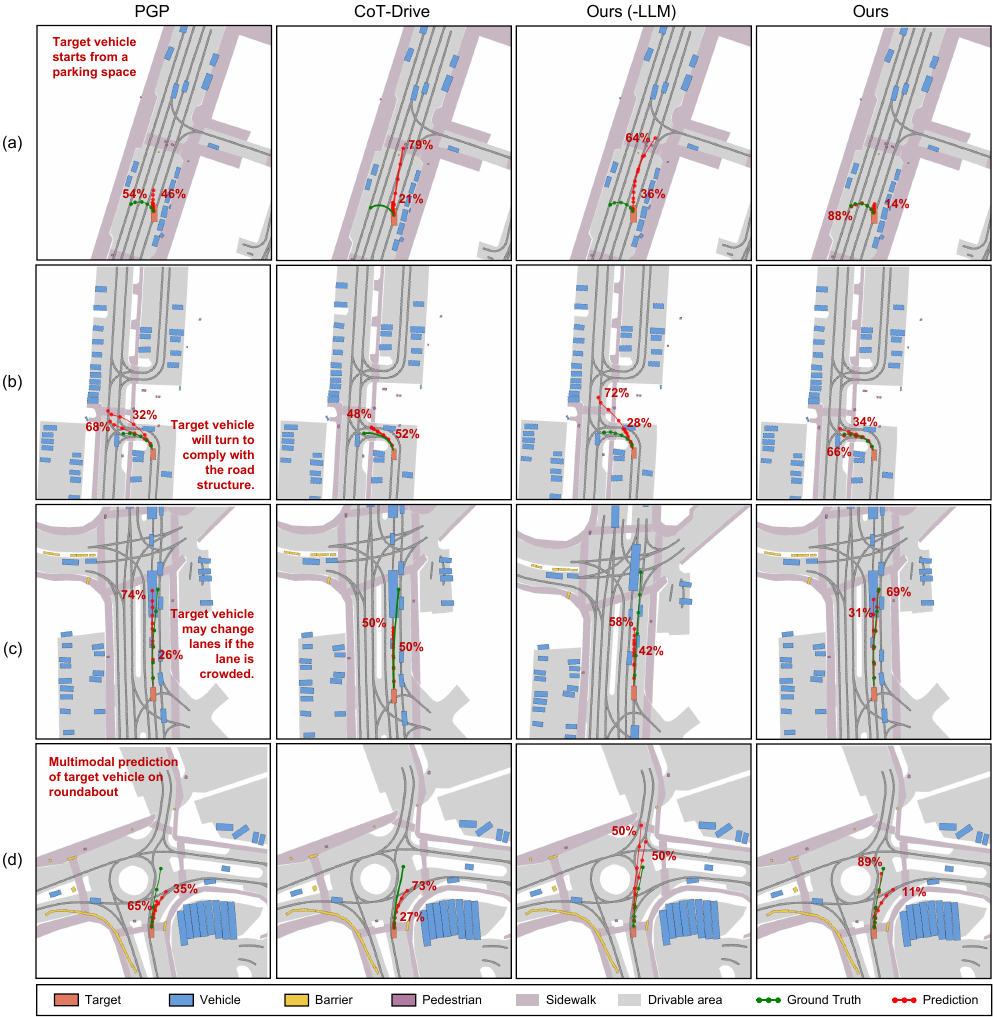}
      \caption{{  Qualitative results of predicted trajectories in corner-case scenarios from the nuScenes-corner dataset, including turning (a), U-turn (b), lane changing (c), and roundabout (d) challenging scenes. For each scenario, we visualize the environment around the target vehicle, along with the {{predicted trajectories}} and the {ground truth}. For clarity, the number of predicted trajectories is set to \(k = 2\), with the probability of predictions displayed in the subplots. "-LLM" represents the model without the Language-Enhanced Encoder in the Memory Module.}}
  \label{visualization-nuScenes}
\end{figure*}

\begin{figure*}
  \centering
  \includegraphics[width=0.95\linewidth]{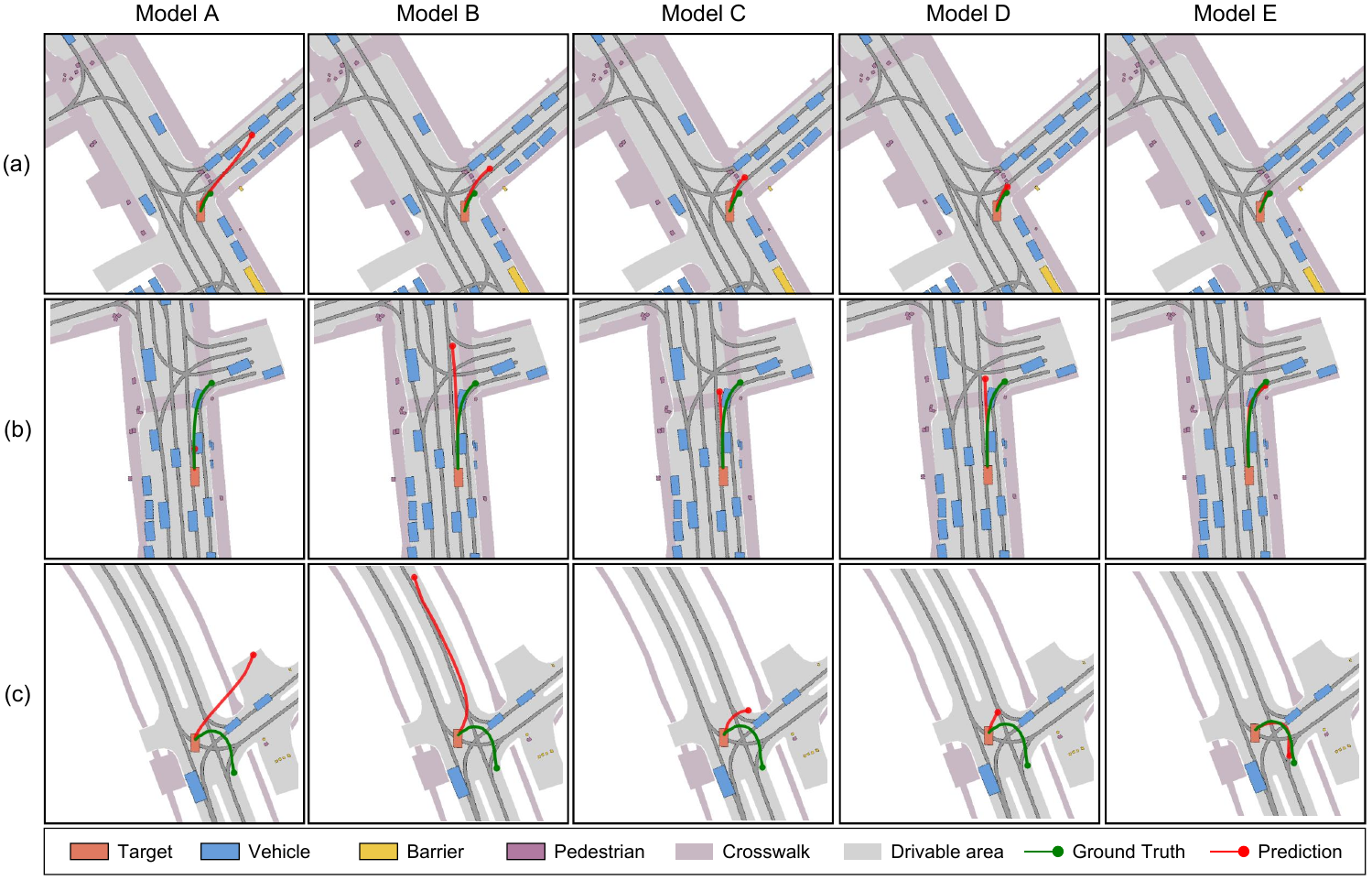}
  \caption{ Qualitative comparisons of the most probable predictions for different ablation models on the nuScenes dataset are presented. Panel (a) illustrates a corner scenario in which a vehicle waits for pedestrians to cross at a zebra crossing. In this context, our comprehensive model E maintains a motion forecasting that aligns closely with the ground truth. In contrast, other models lacking essential components exhibit significant predictive errors. Panel (b) presents a corner case involving lane changes, where only our complete model delivers accurate predictions. Panel (c) highlights the critical roles of the MoE Decoder and Language-Enhanced Encoder in discerning the driver's true intentions.}
  \label{appendix-nuscenes}
\end{figure*}
\subsection{Qualitative Results}
\subsubsection{Qualitative Results in Complex Urban Scenes}

Figure \ref{visualization-nuScenes} presents a comparative visualization of our proposed model against PGP \cite{deo2021multimodal} (non-LLM-based model) and CoT-Drive \cite{liao2025cot} (LLM-based model) across four challenging scenarios. Scenario (a) features the target agent (orange) initiating movement from a stop line. This scenario is rare and involves significant uncertainty since the vehicle has been stationary, complicating the ability of models to predict its intention. Accurately forecasting its future trajectory necessitates hierarchical modeling of the spatial context and temporal historical states. Our model evidently excels at capturing the target agent's intended motion. Moreover, scenario (b) involves a U-turn, where our model forecasts a 180-degree turn while remaining in the current lane, adhering to the existing road layout and traffic conditions. The trajectory predicted by our model is much closer to the ground truth (green) than the baseline model. In addition, Figure \ref{visualization-nuScenes} (c) showcases an example of lane changing on a crowded road. Here, the model must accurately represent the interactions between the target agent and its surrounding agents to determine the agent’s intention and timing for the lane change. Remarkably, our model effectively detects the target vehicle's plan to change to a less congested lane for acceleration. Finally, in scenario (d), the target agent is positioned in the right-turn lane of a roundabout, illustrating another challenging context. Our model successfully predicts a multimodal distribution of possible maneuvers—either continuing straight or turning right—whereas PGP exclusively detects the right-turn intent, and CoT-Drive fails to recognize the straight maneuver, instead identifying a stop maneuver. Across these challenging scenarios, the predicted trajectories of our model closely match the ground truth in both maneuvering decisions and vehicle positioning, indicating its superior adaptability in urban scenes.

Compared to CoT-Drive, another LLM-based approach, our model yields trajectories that more accurately reflect human intent across these complex scenarios. This superior performance highlights the synergistic potential of integrating MoE with LLMs, leveraging the complementary strengths of specialized expert modules.

\subsubsection{Qualitative Results for Ablation Models} 
\label{Ablation Models}
To more intuitively illustrate the impact of each module in our model on the final prediction results, we conduct a visual analysis of various ablation models using the nuScenes dataset. As shown in Figure \ref{appendix-nuscenes}, in scenario (a), the target agent is waiting at the crosswalk for pedestrians to pass, which is a relatively uncommon situation. Models A and B, lacking the MoE Decoder and LLM enhanced module, fail to effectively capture this crucial context, leading them to incorrectly predict that the target agent will accelerate through the crosswalk. In contrast, the other models successfully incorporate the interactions between the target agent, the pedestrian, and the surrounding environment, with the complete model E yielding the most precise prediction results. Scenario (b) illustrates an instance of lane changing on the road. In this context, the model must infer the target agent's intention to turn right by analyzing the historical state and the current traffic conditions. It is only the complete model E that successfully captures the expected action of the target agent and produces accurate predictions. In scenario (c), the slow-moving target agent is situated at a T-junction, where the sparse context presents challenges for accurate prediction. Notably, only model C, which lacks BEV fusion, and the complete model E, effectively capture the agent's potential intention to make a U-turn. The complete model E benefits from the synergistic interaction of its various modules, enabling superior hierarchical modeling of spatial and temporal factors, thus resulting in predictions that align closely with the current road structure and are more consistent with the ground truth.

\begin{figure*}
  \centering
  \includegraphics[width=0.7\linewidth]{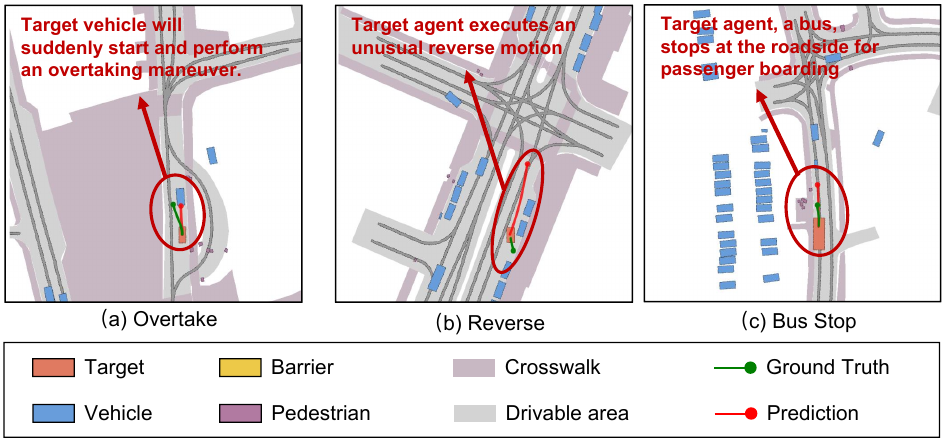}
\caption{{ Qualitative failure cases of WM-MoE on nuScenes-corner dataset. We showcase three real-world driving scenes: (a) Overtake, (b) Reverse, and (c) Bus stop, where the predicted trajectory (red) deviates from the ground truth (green). The target agent is highlighted in orange, and other agents and map elements follow the legend. In panel (a), the target agent accelerates from rest and overtakes; in panel (b), it reverses; in panel (c), a bus pulls over for passenger boarding.}}
  \label{failure-case}
\end{figure*}

{ 
\subsection{Qualitative Failure Analysis and Discussion}
Figure~\ref{failure-case} illustrates long-tail scenarios that remain challenging for WM-MoE. In scenarios (a), the target agent switches from a full stop to an overtake. This abrupt change of intent creates a distribution shift between the observed history and the near future, and the model underestimates the acceleration and interaction required to merge, pass, and re-merge. Moreover, scenarios (b) and (c) involve a reverse maneuver and a bus stopping for passenger boarding. Both require careful reading of weak cues from the scene, including lane geometry, drivable boundaries, and the behavior of nearby agents. The model does not fully infer why the target slows, backs up, or pulls over, and it defaults to common forward driving patterns, resulting in trajectories that are plausible but incorrect.

These challenging cases highlight two main challenges. First, ambiguous or non-informative motion histories make it difficult for the model to disambiguate future intent, particularly in the absence of clear interactions or signaling. Second, the model lacks a robust mechanism to reason about causality in the scene, such as understanding how the presence of pedestrians, barriers, or traffic patterns might influence the target agent’s behavior. While WM-MoE handles the vast majority of corner cases effectively, the failure analyses highlight limitations in extremely atypical, highly uncertain interactions. A promising direction is to narrow these gaps by incorporating change-point detection and intent recognition, enforcing stronger priors conditioned on map structure, and adopting training objectives that promote causal reasoning with explicit counterfactual consistency checks. We expect these additions to calibrate uncertainty when appropriate and to improve forecast quality in rare, safety-critical driving scenes.}

\section{Conclusion}\label{Conclusion}
{ 
This study presents WM-MoE, a world model-based framework designed to address the critical challenge of motion forecasting in rare, safety-critical corner cases for autonomous driving
WM-MoE unifies perception, temporal memory, and decision modules and explicitly addresses real-world data imbalance through a divide-and-conquer Mixture-of-Experts network that allocates specialized capacity to distinct interaction regimes. This allows experts to learn the nuanced dynamics of underrepresented cases without disrupting common patterns. To strengthen scene understanding and long-horizon reasoning, we introduce a lightweight temporal tokenizer that projects trajectories and context into a frozen LLM feature space, injecting commonsense priors without additional LLM training and keeping computation within real-time budgets.
This approach not only improves the model's ability to capture temporal dependencies in trajectories but also enhances training efficiency and prediction accuracy. 
Extensive experiments on four real-world datasets (nuScenes, NGSIM, HighD, and MoCAD) show that WM-MoE achieves SOTA performance compared to existing advanced methods, while also providing faster inference speed. Even in scenarios involving missing data and corner cases, WM-MoE maintains competitive performance, highlighting its strong adaptability and robustness. Additionally, the introduction of the nuScenes-corner dataset, as an additional contribution of this study, focuses specifically on corner-case scenarios to evaluate model robustness under rare and imbalanced real-world driving scenes. Nevertheless, several limitations remain. First, our use of a relatively lightweight LLM is a trade-off between efficiency and reasoning capacity; more powerful models may further improve performance but pose challenges for real-time deployment. Second, while WM-MoE generalizes well across multiple datasets, its performance in highly unstructured or out-of-distribution environments (rural or mixed-modal traffic) remains underexplored. Third, the current framework relies on offline training and inference; although efficient, the LLM module still adds latency and memory relative to purely feedforward backbones. In the future, we plan to expand the nuScenes-corner dataset to include additional cities, diverse weather and nighttime conditions, and a broader range of corner-case scenarios, along with releasing standardized benchmarking protocols. Furthermore, we aim to extend WM-MoE with online adaptation and active learning capabilities, incorporate richer sensor modalities such as LiDAR and radar, and explore the integration of more powerful foundation models to enhance semantic reasoning and improve generalization under limited supervision. In summary, WM-MoE is the first brain-inspired world model compatible with existing motion forecasting models. It significantly enhances robustness and generalization in corner-case scenes, laying a solid foundation for fully AD.}

\printcredits

\section*{Acknowledgements}
This work was supported by the Science and Technology Development Fund of Macau [0122/2024/RIB2, 0215/2024/AGJ, 001/2024/SKL], the Research Services and Knowledge Transfer Office, University of Macau [SRG2023-00037-IOTSC, MYRG-GRG2024-00284-IOTSC], the Shenzhen-Hong Kong-Macau Science and Technology Program Category C [SGDX20230821095159012],   National Natural Science Foundation of China [Grants 52572354], the State Key Lab of Intelligent Transportation System [2024-B001], and the Jiangsu Provincial Science and Technology Program [BZ2024055].

\bibliographystyle{model1_num_names}
\bibliography{cas_refs}
\end{sloppypar}
\end{document}